\documentclass[lettersize,journal]{IEEEtran}

\usepackage{amsmath,amsfonts,amssymb}
\usepackage{algorithm}
\usepackage{algorithmic}
\usepackage{array}
\usepackage[caption=false,font=normalsize,labelfont=sf,textfont=sf]{subfig}
\usepackage{textcomp}
\usepackage{stfloats}
\usepackage{url}
\usepackage{graphicx}
\usepackage{cite}
\usepackage{booktabs}
\usepackage{multirow}
\usepackage{pifont}
\usepackage{color}
\usepackage[table]{xcolor}
\usepackage[capitalize,noabbrev]{cleveref}

\begin{document}

\title{SatSAM2: Motion-Constrained Video Object Tracking in Satellite Imagery using Promptable SAM2 and Kalman Priors}

\author{Ruijie~Fan,
        Junyan~Ye,
        Huan~Chen,
        Zilong~Huang,
        Xiaolei~Wang,
        and~Weijia~Li%
\thanks{R. Fan is with Tsinghua Shenzhen International Graduate School, Tsinghua University, Shenzhen 518055, China, and also with the School of Geospatial Engineering and Science, Sun Yat-sen University, Zhuhai 519082, China.}%
\thanks{J. Ye, H. Chen, Z. Huang, and X. Wang are with the School of Geospatial Engineering and Science, Sun Yat-sen University, Zhuhai 519082, China.}%
\thanks{W. Li is with Tsinghua Shenzhen International Graduate School, Tsinghua University, Shenzhen 518055, China (e-mail: liweijia@sz.tsinghua.edu.cn).}%
\thanks{R. Fan and J. Ye contributed equally to this work.}%
\thanks{Corresponding author: W. Li.}%
}

\markboth{}
{Fan \MakeLowercase{\textit{et al.}}: SatSAM2: Motion-Constrained Video Object Tracking in Satellite Imagery}

\maketitle


\begin{abstract}
Existing satellite video tracking methods often struggle to generalize across diverse scenes, typically requiring scenario-specific training to achieve satisfactory performance, and remain prone to track loss under occlusion and appearance ambiguity. To address these challenges, we propose \textbf{SatSAM2}, a zero-shot satellite video object tracker built upon the foundation model SAM2, designed to adapt promptable segmentation models to the remote sensing domain. SatSAM2 introduces two key components: a \textbf{Kalman Filter-based Constrained Motion Module (KFCMM)} that exploits temporal motion cues and historical trajectory information to stabilize localization and suppress drift, and a \textbf{Motion-Constrained State Machine (MCSM)} that dynamically regulates the tracking process by switching between multiple states according to segmentation confidence and motion consistency. By integrating motion priors with foundation-model-based visual cues, the proposed framework effectively handles challenges in satellite imagery, including small object size, visually similar targets, and frequent occlusions. To support large-scale evaluation, we further construct \textbf{MatrixCity Video Object Tracking (MVOT)}, a synthetic benchmark containing more than 1,500 sequences and 157K annotated frames with diverse viewpoints, illumination conditions, and occlusion scenarios. Extensive experiments on two satellite tracking benchmarks and MVOT demonstrate that SatSAM2 consistently outperforms both conventional trackers and recent foundation-model-based methods, including SAM2 and its variants. Notably, on the OOTB dataset, SatSAM2 achieves a 5.84\% AUC improvement over state-of-the-art approaches. Our code and dataset will be publicly released to encourage further research.
\end{abstract}


\section{Introduction}
\label{sec:intro}

\begin{figure}[t]
  \centering
  \includegraphics[width=\linewidth]{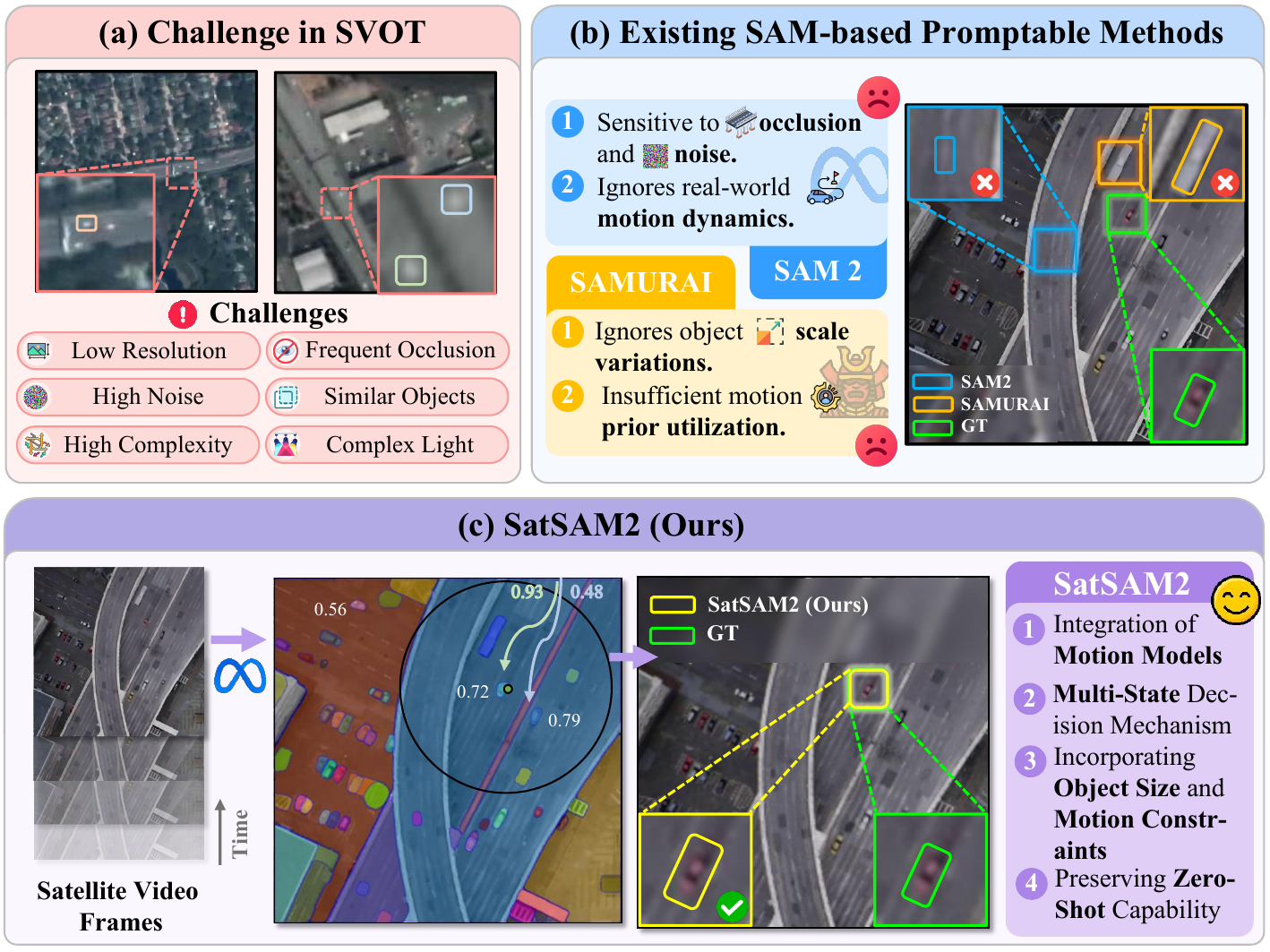}
  \caption{Illustration of \textbf{S}atellite \textbf{V}ideo \textbf{O}bject \textbf{T}racking (SVOT). (a)~Challenges in satellite-based tracking tasks. (b)~Existing promptable methods either lack motion modeling or fail to account for the complete tracking pipeline. (c)~Our approach integrates a Kalman-based motion model with a motion-constrained state machine to enable stable tracking.}
  \label{fig:flag}
\end{figure}

\IEEEPARstart{W}{ith} the advancement of satellite observation technology and the increasing resolution of \textbf{S}atellite \textbf{V}ideo (SV), large-scale and continuous ground object detection and tracking have become feasible~\cite{wang2024satellite}. Compared to traditional moving platforms such as \textbf{C}losed-\textbf{c}ircuit \textbf{T}ele\textbf{v}ision (CCTV), aircraft, and \textbf{U}nmanned \textbf{A}erial \textbf{V}ehicles (UAVs), remote sensing satellite videos offer spatiotemporal monitoring over vast areas, making them valuable in applications such as intelligent transportation systems, digital cities, and urban scene understanding~\cite{yin2021detecting, zhou2019anomalynet, Li_2023_CVPR, Ye_2024_CVPR, Li_2024_CVPR}.  However, despite recent progress in satellite-based scene understanding and object recognition~\cite{LI202326, Li_2023_CVPR, 10478083}, satellite videos typically suffer from low spatial resolution and wide coverage~\cite{10387229}, where targets appear extremely small~\cite{9989433, 10767172} and numerous visually similar objects coexist. Moreover, the presence of atmospheric disturbances and other inevitable imaging effects introduces considerable noise, which blurs the boundaries between objects and their backgrounds. Occlusions caused by overhead structures (e.g., bridges) and complex illumination effects (e.g., shadows from tall buildings or varying surface reflectance) often hinder the cross-domain generalization of traditional fully supervised visual trackers, as illustrated in \cref{fig:flag}(a). Consequently, owing to the substantial domain gap, traditional visual trackers suffer an inevitable and significant performance degradation when directly applied to satellite video tracking tasks~\cite{yang2023track, dendorfer2020mot20, milan2015joint, milan2021motchallenge}.

The \textbf{S}egment \textbf{A}nything \textbf{M}odel (SAM)~\cite{Kirillov_2023_ICCV}, a foundation model for promptable zero-shot segmentation, demonstrates strong generalization across diverse domains. Building on SAM, SAM2~\cite{ravi2025sam} introduces a streaming memory mechanism that supports frame-by-frame video processing with long-range temporal context, offering new opportunities for zero-shot video object tracking. However, directly applying SAM2 to satellite videos remains challenging due to severe occlusion, subtle appearance changes, and the presence of visually similar objects. As illustrated in \cref{fig:flag}(b), SAM2 may lose targets under bridge occlusions or confuse multiple vehicles moving in opposite directions. Moreover, satellite targets are typically rigid with limited scale variation, making Kalman filter-based propagation~\cite{yang2024samuraiadaptingsegmentmodel, 10.1115/1.3662552, welch1995introduction, khodarahmi2023review, kim2018introduction, Grewal2025} unreliable---predicted boxes gradually shrink during partial occlusions, ultimately leading to track loss.

To address these challenges, we propose SatSAM2 (\cref{fig:flag}(c)), a novel framework that integrates SAM2 into satellite video tracking while incorporating motion priors and historical trajectory information. Our design takes advantage of specific constraints in satellite imagery---namely, target rigidity, relatively consistent motion, and minimal scale variation. By leveraging historical motion cues to define state-aware bounding box strategies, SatSAM2 effectively recovers lost targets following temporary occlusions. Furthermore, the rigidity constraint helps suppress bounding box jitter due to motion blur and reduces confusion caused by segmentation errors from SAM2.

We also introduce a large-scale dataset, \textbf{M}atrixCity \textbf{V}ideo \textbf{O}bject \textbf{T}racking (MVOT), derived from the synthetic MatrixCity environment. This dataset includes over 1,500 video sequences and 157,900 annotated frames, representing a significant improvement in scale compared to existing satellite video datasets. MVOT contains sequences across diverse satellite viewing angles, lighting conditions, and occlusion scenarios, providing a robust benchmark for evaluating tracker performance under varying conditions.

Our main contributions are summarized as follows:
\begin{itemize}
    \item We propose SatSAM2, the first SAM2-based satellite video tracking method that introduces a promptable foundation model into remote sensing scenarios. By coupling a motion-constrained state machine with Kalman filtering, SatSAM2 achieves reliable zero-shot tracking.
    \item Our design explicitly exploits key physical properties of satellite imagery, including approximate scale invariance under fixed imaging geometry, smooth and near-linear target motion, and the absence of abrupt camera-induced transformations. This enables more accurate target dynamics modeling and improves robustness against occlusion and appearance ambiguity.
    \item We construct MVOT, a comprehensive synthetic dataset for satellite video tracking. It features multi-scenario, multi-view, and multi-illumination conditions, with significantly more sequences than existing datasets. This contribution helps mitigate the issues of limited coverage and inconsistent quality in current SV tracking benchmarks.
    \item Our proposed SatSAM2 consistently outperforms conventional visual trackers, the original SAM2, and its improved variants across various evaluation metrics on multiple real-world and synthetic benchmarks.
\end{itemize}

\section{Related Work}
\label{sec:related_work}

\begin{figure*}[t]
\centering
\includegraphics[width=\textwidth]{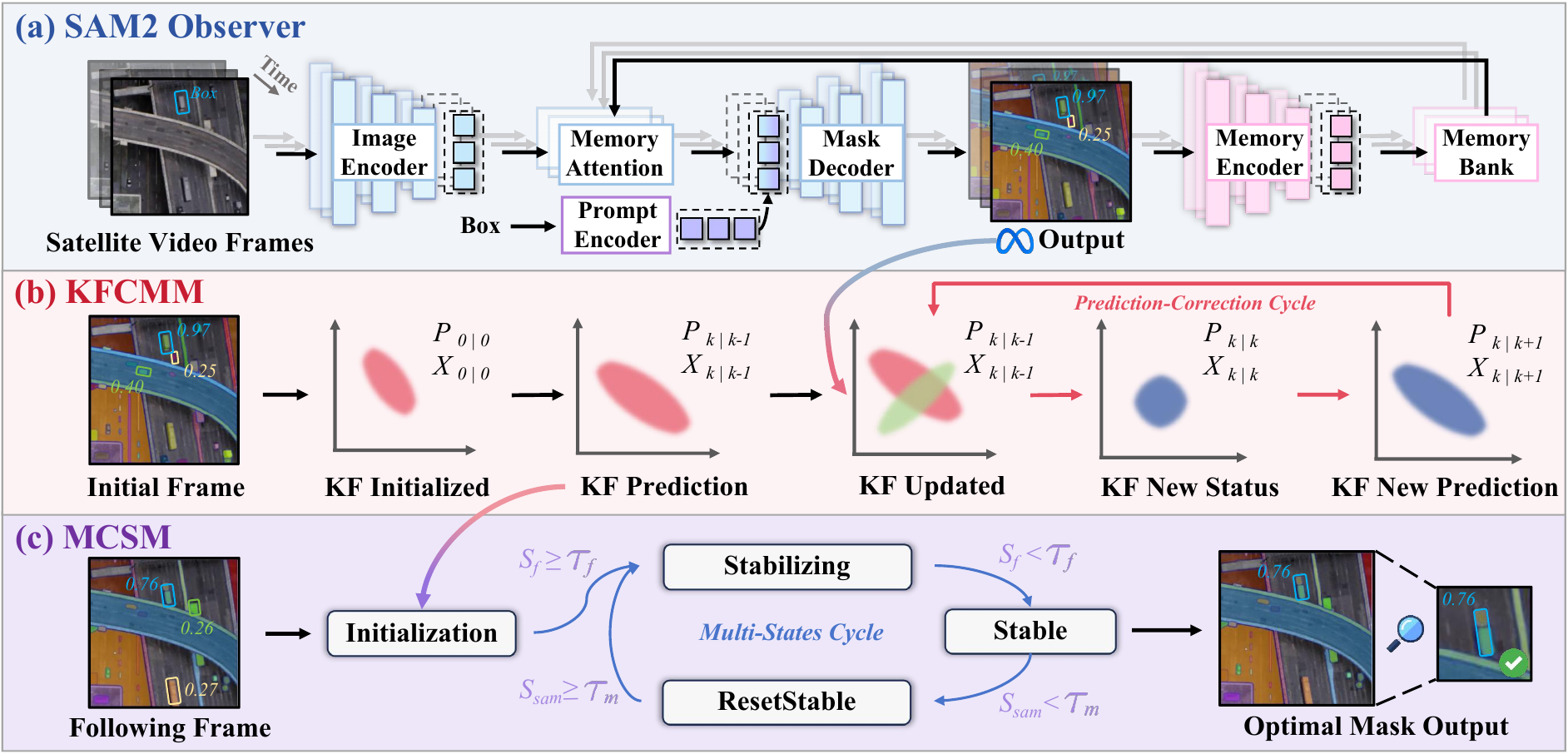}
\caption{\textbf{Overview of the proposed SatSAM2 framework.}
(a)~SAM2 Observer encodes each frame and retrieves candidate masks via memory matching.
(b)~Kalman Filter-based Constrained Motion Model (KFCMM) estimates target dynamics and provides predictive guidance under occlusion.
(c)~Motion-Constrained State Machine (MCSM) adaptively switches between tracking modes based on segmentation confidence and motion consistency.}
\label{fig:pipeline}
\end{figure*}

\subsection{Supervised Object Tracking}
Over the past decade, supervised Object Tracking has progressed from early discriminative models (e.g., DiMP~\cite{Bhat_2019_ICCV}, PrDiMP~\cite{Danelljan_2020_CVPR}) and Siamese-based methods (e.g., LSSiam~\cite{8935538}, SiamCAN~\cite{9369121}) to transformer-based frameworks (e.g., ToMP~\cite{Mayer_2022_CVPR}, SRTrack~\cite{10359476}, AQATrack~\cite{xie2024autoregressive}) and efficient adaptation methods like LoRAT~\cite{lorat}. While these trackers achieve strong performance on standard benchmarks, they struggle in satellite scenarios due to small targets, low frame rates, occlusion, and drastic viewpoint shifts. To mitigate this, specialized methods have emerged: HRSiam~\cite{9350236} preserves spatial resolution for small-object localization, CRAM~\cite{9875357} incorporates motion priors via Kalman filtering, MCTracker~\cite{yin2021detecting} enhances association using multi-scale temporal cues, and recent work~\cite{10670064} models multiple spatial-temporal relations for robust tracking. Recent remote-sensing-oriented approaches, such as DFTrack~\cite{9803284} and TATrack~\cite{10342836}, further tailor correlation-flow modeling and target-aware transformer designs to satellite-video characteristics. Additionally, recent advances have explored feature enhancement with multi-level matching~\cite{Yang2023SVOT}, as well as integrating temporal information and trajectory prediction (e.g., SiamTITP~\cite{11051136}) to further stabilize tracking under complex dynamics. These adaptations highlight the domain-specific challenges of remote sensing tracking, where general trackers often fail to generalize.

Our approach is grounded in zero-shot learning, eliminating the need for task- or dataset-specific pretraining. By directly leveraging the strong generalization capability of SAM2 in conjunction with motion models tailored to remote sensing video, our method achieves effective target tracking in complex and highly challenging scenarios. Although such zero-shot tracking presents inherent difficulties, it demonstrates superior practicality compared with previous approaches and exhibits enhanced adaptability across a broader range of remote sensing scenes.

\subsection{Zero-shot Object Tracking}

Zero-shot object tracking has been enabled by foundation models~\cite{10839471, 11082481, 11087517} such as SAM~\cite{Kirillov_2023_ICCV}, which introduced large-scale promptable segmentation, and SAM2~\cite{ravi2025sam}, which extends this capability to video via temporal memory. SAMURAI~\cite{yang2024samuraiadaptingsegmentmodel} improves zero-shot tracking stability through Kalman filter-based memory selection. In contrast, ROS-SAM~\cite{Shan_2025_CVPR} adapts SAM to remote sensing using LoRA-based fine-tuning and enhanced feature modeling, achieving high-quality interactive segmentation and strong zero-shot performance on moving-object segmentation. Meanwhile, sparsely-supervised~\cite{10541917} and detection-based multi-object approaches~\cite{10571840} have demonstrated that trackers can be effective with reduced or no task-specific annotations. Despite these advances, zero-shot tracking in satellite videos remains challenged by domain shifts, mask drift, and severe occlusions.

Our method introduces a multi-stage filtering mechanism that models and optimizes the tracking process by incorporating motion patterns unique to remote sensing satellite videos. This design enables the model to effectively exploit key information within video sequences while suppressing the influence of noisy signals on tracking performance. We further demonstrate the effectiveness of our approach through comprehensive ablation studies, which validate the contribution of each component to overall performance.

\subsection{Remote Sensing Tracking Datasets}
Remote sensing tracking datasets serve as fundamental benchmarks for evaluating model performance under satellite-specific conditions. SatSOT~\cite{9672083} pioneered remote SOT benchmarks with 105 HBB-labeled sequences across four object types. VISO~\cite{9625976} scaled this to 3,711 trajectories, focusing on urban scenes. SAT-MTB~\cite{10130311} added multi-task annotations but lacked OBBs, which are important for orientation modeling. OOTB~\cite{CHEN2024212} incorporated OBBs but reused many existing sequences. AIR-MOT~\cite{10130311} is a multi-object, multi-task benchmark that follows a different evaluation protocol, making it less directly comparable to single-object tracking methods. Overall, most datasets are limited in length, diversity, and annotation completeness, especially for small or low-contrast objects. This hinders fair evaluation under real-world remote sensing conditions, highlighting the need for longer, more diverse, and richly annotated benchmarks covering SOT and MOT. To address these limitations, our proposed MVOT dataset provides 3--10$\times$ more frames than existing benchmarks, together with substantially richer variations in viewing angles and illumination conditions.

\section{Method}
\label{sec:method}

As illustrated in \cref{fig:pipeline}, this paper presents an innovative zero-shot satellite video object tracker named \textbf{SatSAM2}, which is built upon a motion-constrained state machine architecture. The overall framework is composed of three main modules: the SAM2 Observer, the \textbf{K}alman \textbf{F}ilter-based \textbf{C}onstrained \textbf{M}otion \textbf{M}odule (KFCMM), and the \textbf{M}otion-\textbf{C}onstrained \textbf{S}tate \textbf{M}achine (MCSM).
The SAM2 Observer extracts multi-scale visual features from satellite remote sensing videos via an image encoder, and fuses them with long-term contextual information stored in a memory bank through a memory attention mechanism. Conditioned on box prompts, candidate segmentation masks are generated using the SAM2 mask decoder.
The KFCMM initializes the motion model using the initial box prompt and continually updates the state space with high-confidence bounding boxes observed by the SAM2 Observer, thereby refining the target's motion estimation over time.
Finally, the MCSM governs the tracking process by dynamically choosing estimation strategies according to the current tracking status, enabling SatSAM2 to maintain robust and accurate target localization across varying conditions.

Rather than simply attaching a classical motion model to SAM2, SatSAM2 explicitly encodes physical assumptions that are characteristic of satellite video: (1)~approximate scale invariance under fixed imaging geometry, (2)~smooth and near-linear target motion in the image plane, and (3)~the absence of abrupt camera-induced transformations. These domain-specific priors are embedded into the KFCMM (via constrained state transitions and area preservation) and the MCSM (via motion-consistency-driven state logic), which jointly regulate mask selection and state transitions. This design distinguishes SatSAM2 from naive SAM2+Kalman filter combinations, which lack such structured integration and fail to exploit the unique regularity of satellite imagery (see \cref{tab:ablation} for empirical validation).

\subsection{SAM2 Observer}

\paragraph{Satellite Video Feature Extraction}
To continuously track a target, visual trackers typically need to extract target-relevant information from each video frame. We employ the SAM2 video segmentation model as the feature extraction component of our tracker. Each frame of the satellite video is passed into the Hiera image encoder~\cite{ryali2023hiera}, which is pretrained with MAE~\cite{He_2022_CVPR}, to obtain unconditional embeddings. These image embeddings are then refined through cross-attention with past frame features stored in a memory bank, along with any newly provided prompts, via a memory-attention module.

\paragraph{Memory Selection}
Prolonged occlusions in satellite video tracking can easily contaminate the memory bank with irrelevant information. To mitigate this issue, we adopt the memory selection strategy proposed in SAMURAI. Specifically, for each frame \( i \), a memory candidate \( m_i \) is only selected when the following three criteria are simultaneously satisfied:
\begin{equation}
s^{(i)}_{\text{mask}} > \tau_{\text{mask}}, \quad s^{(i)}_{\text{obj}} > \tau_{\text{obj}}, \quad s^{(i)}_{\text{kf}} > \tau_{\text{kf}},
\label{eq:selection}
\end{equation}
where \( s^{(i)}_{\text{mask}} \), \( s^{(i)}_{\text{obj}} \), and \( s^{(i)}_{\text{kf}} \) denote the mask affinity score, the object existence score (from SAM2), and the motion consistency score (from the Kalman filter), respectively. The corresponding thresholds are \( \tau_{\text{mask}}, \tau_{\text{obj}}, \tau_{\text{kf}} \).

For each current frame \( t \), the memory bank \( \mathcal{M}_t \) consists of at most \( N_{\max} \) valid memory entries selected in reverse chronological order from past frames that satisfy the condition in \cref{eq:selection}:
\begin{equation}
\mathcal{M}_t = \left\{ m_i \;\middle|\; i < t,\; m_i \text{ satisfies Eq.~\eqref{eq:selection}} \right\}, \quad |\mathcal{M}_t| \leq N_{\max}.
\end{equation}
If fewer than \( N_{\max} \) valid entries are available, all qualifying entries are used.

After obtaining the informative memory set \( \mathcal{M}_t \), the current frame features \( f_t \), along with the sparse box prompts \( p_t \) from the prompt encoder, are fed into the mask decoder. The decoder produces three candidate masks, and the selected mask features are subsequently encoded by the memory encoder and stored into the memory bank.


\paragraph{Relationship to SAMURAI}
Although our SAM2 Observer shares the memory selection strategy with SAMURAI~\cite{yang2024samuraiadaptingsegmentmodel}, the two systems differ fundamentally. SAMURAI employs only a standard linear Kalman filter for scoring candidate masks during memory selection, without any motion-constrained state management. In contrast, SatSAM2 introduces a dedicated motion-constrained Kalman filtering module (KFCMM) that encodes satellite-specific physical constraints, together with a novel motion-constrained state machine (MCSM) that governs the full tracking pipeline.
The target is initialized via a bounding box prompt provided in the first frame. In subsequent frames, the memory bank retains at most $N_{\max}$ high-confidence target states from prior frames, and the prompt encoder conditions mask decoding on the latest estimated bounding box.

\begin{algorithm}[tb]
\caption{Decision Logic in \textit{Stable} State}
\label{alg:stable_state}
\textbf{Input}: $s_{\text{sam}}, s_{\text{kf}}$\\
\textbf{Output}: Target mask $M$, Kalman update

\begin{algorithmic}[1]
\IF{$s_{\text{sam}} > \tau_h$}
    \STATE $M \gets M_{\text{sam}}$
    \STATE Update Kalman with $M$
\ELSIF{$\tau_m < s_{\text{sam}} \leq \tau_h$}
    \STATE $M \gets M_{\text{sam}}$
    \STATE Kalman: \textbf{skip update}, propagate state
\ELSIF{$s_{\text{sam}} \leq \tau_m \land s_{\text{kf}} > \tau_{\text{kf}}$}
    \STATE $M \gets \arg\max_i \text{IoU}(M_i, \hat{M}_{\text{kf}})$
    \STATE Update Kalman with $M$
\ELSE
    \STATE Kalman: \textbf{skip update}
    \STATE $s_f \gets 0$
    \STATE Switch state: \textit{ResetStable}
\ENDIF
\end{algorithmic}
\end{algorithm}

\subsection{Kalman Filter-based Constrained Motion Module}

\paragraph{Kalman State Vector Construction}
Traditional tracking methods typically learn the target model by minimizing a discriminative objective function to localize the target in each frame, which makes it difficult to model the motion state of the target itself. In SatSAM2, we incorporate a linear Kalman filter into the tracking module to progressively model the target's motion state by continuously assimilating high-confidence observations from the SAM2 model. The Kalman state vector is defined as follows:
\begin{equation}
\mathbf{x} = [x, y, w, h, \dot{x}, \dot{y}, \dot{w}, \dot{h}, S]^\top,
\end{equation}
where \( x \) and \( y \) denote the center coordinates of the bounding box (bbox), while \( w \) and \( h \) represent its width and height, respectively. Their first-order derivatives (denoted by dots) form the next four elements in the state vector. These quantities represent the variables to be estimated by the Kalman filter, from which prior estimates of the target bounding box can be derived in subsequent Kalman iterations.

The last element \( S \) represents the object area, estimated from the mask obtained via the initial prompt box segmentation. Under the assumption of near-orthographic satellite projection and constrained object motion, we treat \( S \) as a constant during subsequent Kalman filter predictions.

\paragraph{Kalman Filter Formulation}
The Kalman filter is an efficient recursive estimator that infers the state of a dynamic system from noisy and incomplete measurements~\cite{10.1115/1.3662552}. In SatSAM2, we deploy a linear Kalman filter tailored with motion constraints specific to remote sensing video tracking~\cite{yang2024samuraiadaptingsegmentmodel}, resulting in the KFCMM. The update process follows the standard prediction-correction equations:
\begin{align}
\hat{x}_{k|k-1} &= A \hat{x}_{k-1|k-1}, \label{eq:predict_state} \\
P_{k|k-1} &= A P_{k-1|k-1} A^T + Q, \label{eq:predict_cov} \\
K_k &= P_{k|k-1} H^T (H P_{k|k-1} H^T + R)^{-1}, \label{eq:kalman_gain} \\
\hat{x}_{k|k} &= \hat{x}_{k|k-1} + K_k (z_k - H \hat{x}_{k|k-1}), \label{eq:update_state} \\
P_{k|k} &= (I - K_k H) P_{k|k-1}, \label{eq:update_cov}
\end{align}
where \cref{eq:predict_state,eq:predict_cov} perform the time update (prediction step), propagating the prior state and covariance using the state transition matrix $A$; \cref{eq:kalman_gain} computes the Kalman gain $K_k$, balancing trust between prediction and new observations; and \cref{eq:update_state,eq:update_cov} refine the state and covariance using the measurement $z_k$.

\paragraph{KFCMM Prediction-Correction Cycle}
In each prediction-correction cycle, the prior state estimate \( \hat{\mathbf{x}}_{t+1|t} \) is given by the state transition equation:
\begin{equation}
\hat{\mathbf{x}}_{t+1|t} = \mathbf{F} \, \hat{\mathbf{x}}_{t|t},
\end{equation}
where \( \mathbf{F} \) is the state transition matrix defined as:
\begin{equation}
\mathbf{F} =
\begin{bmatrix}
\mathbf{I}_4 & \Delta t \cdot \mathbf{I}_4 & \mathbf{0}_{4\times1} \\
\mathbf{0}_{4\times4} & \mathbf{I}_4 & \mathbf{0}_{4\times1} \\
\mathbf{0}_{1\times4} & \mathbf{0}_{1\times4} & 1
\end{bmatrix}
\end{equation}
with \( \mathbf{I}_4 \) denoting a \( 4 \times 4 \) identity matrix, and \( \mathbf{0}_{m \times n} \) representing a zero matrix of shape \( m \times n \).

\paragraph{Prior Estimation Selection in KFCMM}
For each candidate mask \( M_j \) selected by the SAM2 Observer, let \( S_M^j \) be its area and \( (x_M^j, y_M^j) \) its center. We define a Kalman-based motion score \( s_{\text{kf}}^j \) that quantifies the motion consistency of the mask as:
\begin{equation}
s_{\text{kf}}^j =
\begin{cases}
\text{IoU}(\hat{\mathbf{x}}_{t+1|t}, M_j), & \text{if } \frac{S}{S_M^j} \in \mathcal{D}, \\
0, & \text{otherwise},
\end{cases}
\label{eq:skf}
\end{equation}
where \( \text{IoU}(\cdot) \) denotes the Intersection-over-Union between the Kalman-predicted bbox and the candidate mask \( M_j \), and \( \mathcal{D} \subset \mathbb{R} \) is the acceptable deformation range. In our implementation, we set \( \mathcal{D} = [0.5, 2.0] \).
This range is empirically determined and reflects the observation that, in satellite imagery, target area variation is typically bounded: objects rarely shrink below half or expand beyond twice their initial size due to the near-orthographic projection and limited altitude change. We further show in \cref{fig:sensitivity} that performance is not sensitive to this parameter within a reasonable range.

In addition, we compute the Euclidean distance \( d_j \) between the predicted bbox center and the center of mask \( M_j \). Let \( \alpha_{\text{kf}} \in [0,1] \) be a weighting factor controlling the influence of the motion model, and \( s_{\text{sam}}^j \) be the affinity score predicted by SAM2. The distance condition is encoded by an indicator function:
\begin{equation}
U_j =
\begin{cases}
1, & \text{if } d_j < d_{\max}, \\
0, & \text{otherwise},
\end{cases}
\end{equation}
where \( d_{\max} = \sqrt{S} \) is set adaptively as the square root of the initial target area \( S \), ensuring the spatial gate scales proportionally to target size.

The final selected mask \( M^* \) is determined by a weighted score that fuses the motion consistency and segmentation confidence:
\begin{equation}
M^* = \arg\max_{M_j} \left( \left( \alpha_{\text{kf}} \cdot s_{\text{kf}}^j + (1 - \alpha_{\text{kf}}) \cdot s_{\text{sam}}^j \right) \cdot U_j \right).
\label{eq:finalscore}
\end{equation}

\subsection{Motion-Constrained State Machine}
Relying solely on a fixed target selection mechanism such as that defined in \cref{eq:finalscore} often fails to fully exploit the rich measurement information provided by SAM2 and the predictive capability of the KFCMM, especially under the unique challenges of satellite video tracking. To address this, we design a motion-constrained state machine for the visual tracker, which governs the integration of segmentation and motion estimation through five well-defined states: \textit{Uninitialized}, \textit{Initialized}, \textit{Stabilizing}, \textit{Stable}, and \textit{ResetStable} (RS).

At the beginning of tracking, the state machine enters the \textit{Uninitialized} state and transitions to \textit{Initialized} by selecting the highest-scoring mask predicted by SAM2 in the first frame. It then enters the \textit{Stabilizing} state, where the tracker accumulates a stability score \( s_f \). If the SAM2 affinity score exceeds a high-confidence threshold \( \tau_h \), the score \( s_f \) is incremented. Once \( s_f \) reaches a frame-wise threshold \( \tau_f \), the tracker transitions to the \textit{Stable} state. Otherwise, \( s_f \) is reset to zero and the stabilization process restarts.

In the \textit{Stable} state, the tracker assumes that the target has been reliably localized and begins making decisions based on both appearance and motion cues. The decision logic is detailed in \cref{alg:stable_state}. Two scores guide this process: the SAM2 affinity score \( s_{\text{sam}} \), which reflects the confidence of the segmentation output, and the Kalman motion score \( s_{\text{kf}} \), which quantifies the agreement between the predicted and observed target positions. Based on predefined thresholds \( \tau_h \), \( \tau_m \), and \( \tau_{\text{kf}} \), the tracker either directly adopts the SAM2 mask \( M_{\text{sam}} \), selects the mask with maximum Intersection-over-Union (IoU) against the Kalman prediction \( \hat{M}_{\text{kf}} \), or skips Kalman updates altogether. When both appearance and motion confidence are low, the stability score \( s_f \) is reset and the tracker transitions to the \textit{ResetStable} state. This mechanism ensures a dynamic balance between visual reliability and motion consistency during long-term tracking.

The \textit{ResetStable} (RS) state serves as a recovery mechanism following degradation from the \textit{Stable} state, typically triggered by occlusion. Unlike the \textit{Stabilizing} state, it conservatively maintains the tracking output using existing historical information, but does not increment the stability score until a high-confidence observation is received again. The RS state is a key component of the MCSM, as it prevents premature re-stabilization under noisy observations and enables the tracker to gracefully bridge occlusion gaps using purely predictive Kalman estimates.

\subsection{Detailed Tracking Pipeline}
\label{sec:pipeline_detail}

We describe the end-to-end tracking procedure involving the KFCMM and MCSM modules across the five tracking states.

The SAM2 Observer reads video frames sequentially and segments the target in each frame based on the initial bounding box prompt. At the first frame, the MCSM selects the most confident mask (with the highest affinity score) among the top three masks produced by the SAM2 Observer and passes it to the KFCMM for initialization.

From the second to the tenth frame, the system enters the \textit{Stabilizing} phase. During this period, as the KFCMM has not yet received sufficient observations, the system relies more on the SAM2 Observer. Masks with confidence scores above a certain threshold are used as valid observations for updating the KFCMM.

Once the accumulated stability score $s_f$ exceeds a predefined threshold $\tau_f$, the MCSM transitions to the \textit{Stable} state. When partial occlusion occurs, the SAM2 Observer may produce all three masks with confidence scores below $\tau_m$, yet one of them may still have a high IoU with the predicted mask from the KFCMM. If this mask achieves a sufficiently high motion consistency score $s_{kf}$, it is selected as the final tracking result and used to update the KFCMM.

The \textit{ResetStable} state is activated when the target is heavily occluded and the SAM2 output becomes unreliable, i.e., the predicted masks have low overlap with the KFCMM prediction. In this case, the system relies solely on the motion prediction from the KFCMM. Since no valid observation is available, the KFCMM skips updates, producing a predictive bounding box that glides over the occlusion region. Once the occlusion ends and a high-confidence mask is detected, the MCSM transitions back to the \textit{Stabilizing} state and the KFCMM resumes updates, ensuring robust tracking continuity.


\section{MVOT Dataset}\label{sec:dataset}

\begin{figure}[t]
  \centering
  \includegraphics[width=0.45\textwidth]{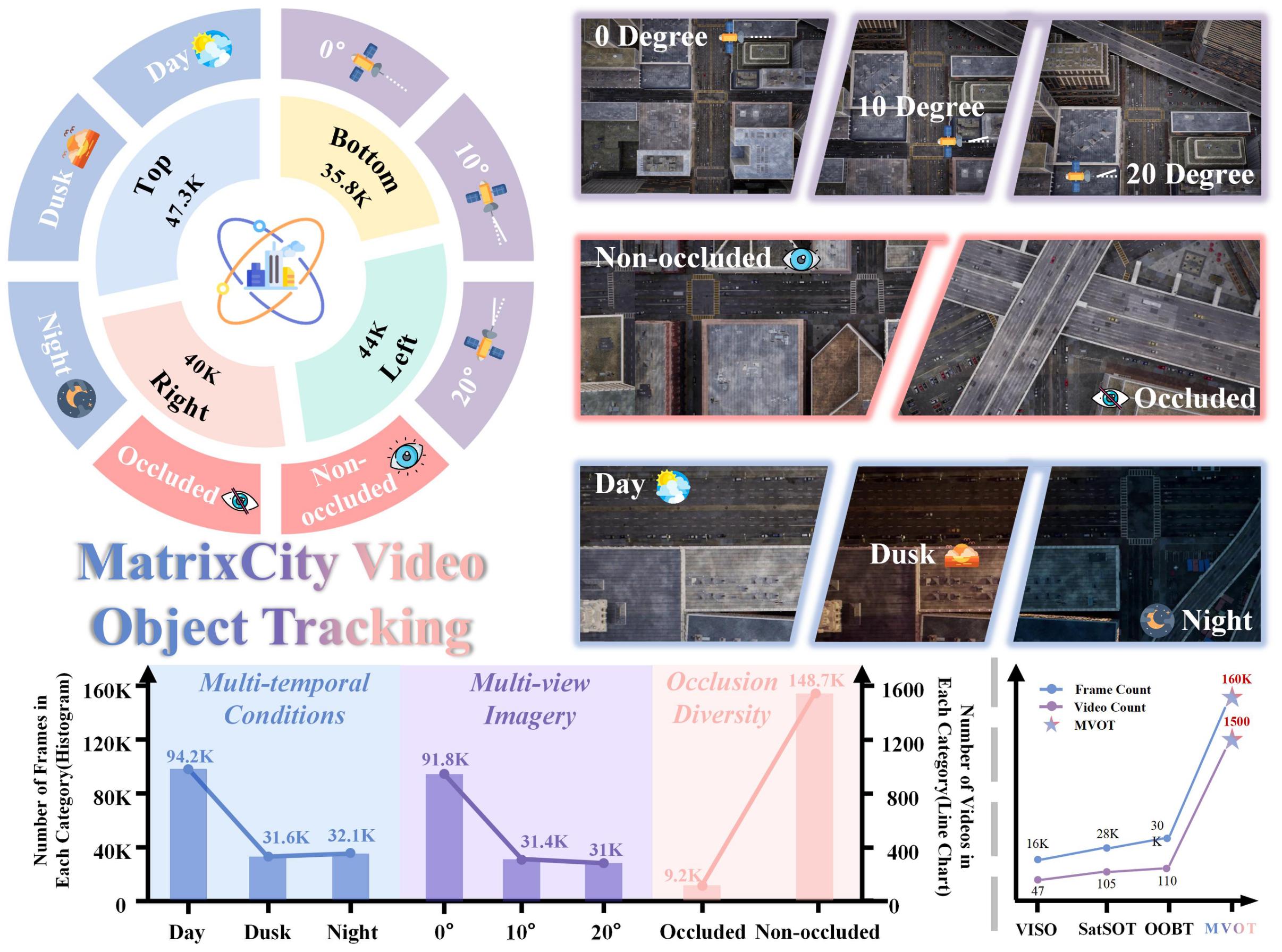}
  \caption{\textbf{Illustration of the MVOT dataset. The MVOT dataset encompasses multi-temporal conditions, including Night, Dusk, and Day; multi-viewpoint settings at $0^\circ$, $10^\circ$, and $20^\circ$; as well as both occluded and non-occluded scenarios. In total, it comprises over 157k frames and more than 1,500 video sequences.}}
  \label{fig:dataset}
\end{figure}

We present MatrixCity Video Object Tracking (MVOT), a remote sensing video object tracking dataset. MVOT is constructed from the synthetic MatrixCity~\cite{li2023matrixcity} environment and designed to evaluate performance under diverse conditions, including multiple scenes and attributes. As illustrated in \cref{fig:dataset}, MVOT comprises approximately 160,000 frames across more than 1,500 video sequences, each accompanied by detailed target annotations, significantly exceeding the scale of existing remote sensing video datasets.
MVOT covers three major dimensions of variation:
\begin{itemize}
    \item \textbf{Illumination conditions}: day, dusk, and night;
    \item \textbf{Viewing angles (zenith angles)}: 0\textdegree, 10\textdegree, and 20\textdegree;
    \item \textbf{Occlusion attributes}: presence or absence of target occlusion within the sequence.
\end{itemize}

Each video sequence in MVOT consists of 100 frames with a spatial resolution of 1024$\times$1024 pixels. The data are collected from four distinct spatial quadrants in the MatrixCity environment, referred to as Right, Left, Bottom, and Top, which correspond to four predefined camera trajectory regions in the official MatrixCity setting~\cite{li2023matrixcity}. Each region represents typical urban scenes rich in tall and low-rise buildings, roads and elevated bridges, ports, and parking lots. This diversity enables comprehensive evaluation across realistic satellite video scenarios.

MVOT is generated using the MatrixCity rendering scripts in Unreal Engine (UE). Camera positions are sampled at regular intervals along predefined trajectories, with a fixed camera height to simulate satellite altitude. Scene pitch angles are varied to produce 0\textdegree, 10\textdegree, and 20\textdegree{} viewing conditions. Illumination is simulated by adjusting sunlight color, intensity, and atmospheric scattering parameters within the UE rendering pipeline. Target bounding box annotations are obtained via the UE actor-tracking API, ensuring pixel-accurate ground truth. The complete data generation pipeline and scripts will be publicly released.

\begin{table}[t]
\centering
\fontsize{9pt}{9pt}\selectfont
\setlength{\tabcolsep}{1mm}
\caption{\textbf{Comparison of statistics and properties between our MVOT dataset with previous datasets.}}
\begin{tabular}{l|cc|ccc}
\toprule
\multirow{2}{*}{Dataset} & \multicolumn{2}{c|}{Count} & \multicolumn{3}{c}{Diversity} \\
 & Video & Frame & Illumination & Angles & Occlusion \\
\midrule
VISO & 47 & 16,204 & \ding{55} & \ding{55} & \ding{55} \\
SatSOT & 105 & 27,664 & \ding{55} & \ding{55} & \ding{55} \\
OOTB & 110 & 29,890 & \ding{55} & \ding{55} & \ding{55} \\
AIR-MOT & 149 & 1,043 & \ding{55} & \ding{55} & \ding{55} \\
SAT-MTB & 249 & 50,046 & \ding{55} & \ding{55} & \ding{55} \\
\rowcolor{gray!15}
\textbf{MVOT} & \textbf{1,500} & \textbf{157,900} & \checkmark & \checkmark & \checkmark \\
\bottomrule
\end{tabular}
\label{tab:dataset_stats}
\end{table}

\subsection{Illumination Conditions}
Conventional remote sensing video datasets typically lack categorization under different illumination conditions, making it difficult to evaluate the robustness of trackers to varying lighting scenarios. Leveraging the large-scale MatrixCity open-source virtual dataset, we utilize the Unreal Engine (UE) to systematically control scene illumination and generate over 90,000 daytime scenes, along with approximately 32,000 scenes for both dusk and nighttime conditions. In the daytime setting, lighting is relatively uniform, and object appearances retain rich details. The dusk scenes are characterized by strong directional lighting, where tall structures cast long shadows on the opposite side, and the overall scene exhibits a warm, golden tone typical of sunset; object details are moderately reduced. In contrast, the nighttime scenes present dim road illumination and prominent artificial lighting on buildings. Object details are significantly diminished, and many targets blend into the background, making them difficult to distinguish---thus posing a substantial challenge to visual tracking algorithms.

\subsection{Viewpoint and Occlusion Attributes}
In addition to illumination, the observation angle of the camera in remote sensing videos often varies, which affects the appearance of targets and introduces additional tracking challenges. To facilitate robustness evaluation under viewpoint changes, we construct three types of scenes with different viewing angles: 0\textdegree{} (approximately 92,000 frames), 10\textdegree, and 20\textdegree{} (approximately 31,000 frames). In the 0\textdegree{} scenes, only the top view of objects is visible, while the 10\textdegree{} and 20\textdegree{} scenes expose more side-view information. At a 20\textdegree{} viewing angle, occlusions become more prominent---tall buildings may block roads, causing targets to disappear into blind zones and reappear on the other side after several frames.
For occlusion-specific evaluation, we also construct a dedicated subset containing sequences with explicit occlusions (approximately 9,200 frames across 92 videos). This enables direct assessment of a tracker's robustness to occluded scenarios.
Furthermore, all the aforementioned attributes (illumination, viewpoint, and occlusion) are combinable. For instance, our dataset includes challenging sequences such as ``20\textdegree-nighttime-occlusion'', allowing for fine-grained analysis of tracker performance under specific compound conditions.

\subsection{Synthetic-to-Real Domain Gap Analysis}
Although MVOT provides a large-scale and systematically controlled benchmark for remote sensing video object tracking, it remains fundamentally different from real satellite imagery. Synthetic data generated from the MatrixCity environment allow precise ground-truth annotations and consistent environmental control but inevitably simplify real-world complexities. In real satellite observations, continuous viewing geometry, varying ground sample distance (GSD), imperfect band alignment, and sensor-level artifacts such as motion blur, haze, and atmospheric interference introduce diverse degradations that are difficult to reproduce synthetically.

To this end, MVOT is designed as a complementary benchmark rather than a substitute for real-world datasets. It enables controlled studies on illumination, viewpoint, and occlusion while supporting scalable pretraining and robustness evaluation. Future extensions will explore domain adaptation and physics-based rendering refinement to further narrow the synthetic-to-real gap and improve transferability to operational satellite tracking scenarios.


\begin{table*}[t]
    \centering
    \caption{\textbf{Detailed hyperparameter settings.} The parameters are organized by module. All values are fixed across OOTB, SatSOT, SAT-MTB, and MVOT.}
    \label{tab:hyperparams}
    \setlength{\tabcolsep}{6pt}
    \renewcommand{\arraystretch}{1.15}
    \begin{tabular}{lcl | lcl}
        \toprule
        \multicolumn{3}{c|}{\textbf{General \& Memory Selection}} & \multicolumn{3}{c}{\textbf{Motion (KFCMM) \& State Machine (MCSM)}} \\
        \cmidrule(r){1-3} \cmidrule(l){4-6}
        \textbf{Param} & \textbf{Value} & \textbf{Description} & \textbf{Param} & \textbf{Value} & \textbf{Description} \\
        \midrule
        Backbone & Hiera-B+ & Pretrained SAM 2.1 model & $\alpha_{kf}$ & 0.2 & Motion weight (vs. visual) \\
        $\tau_{mask}$ & 0.5 & Min. mask affinity for memory & $\mathcal{D}$ & [0.5, 2.0] & Allowed area deformation \\
        $\tau_{obj}$ & 0.0 & Object existence threshold & $d_{max}$ & $\sqrt{S_{init}}$ & Adaptive spatial gate \\
        $\tau_{kf}$ & 0.0 & Motion consistency threshold & $\tau_{h}$ & 0.3 & High-confidence threshold \\
        $N_{max}$ & 16 & Max memory bank size & $\tau_{m}$ & 0.0 & Propagation threshold \\
        Mode & Zero-shot & No weight updates & $T_{f}$ & 12 & Stability duration (frames) \\
        - & - & - & $T_{m}$ & 5 & Reset trigger (frames) \\
        \bottomrule
    \end{tabular}
\end{table*}

\begin{table*}[t]
\centering
\caption{\textbf{Comparison of different trackers on three datasets (OOTB, MVOT, SatSOT) using three metrics: AUC, Precision (P), and Normalized Precision ($P_{\text{norm}}$).} Best results are bolded, second best underlined.}
\begin{tabular}{l|c|ccc|ccc|ccc}
\toprule
\multirow{2}{*}{\textbf{Tracker}} & \multirow{2}{*}{\textbf{Venue}} & \multicolumn{3}{c|}{\textbf{OOTB}} & \multicolumn{3}{c|}{\textbf{MVOT}} & \multicolumn{3}{c}{\textbf{SatSOT}} \\
 & & \textbf{AUC} & \textbf{P} & \textbf{$P_{\text{norm}}$} & \textbf{AUC} & \textbf{P} & \textbf{$P_{\text{norm}}$} & \textbf{AUC} & \textbf{P} & \textbf{$P_{\text{norm}}$} \\
\midrule
\multicolumn{11}{l}{\textit{Supervised methods}} \\
DiMP-50          & CVPR'20 & 46.32 & 80.06 & 78.14 & 64.08 & 94.11 & 93.13 & 46.86 & 57.20 & 46.01 \\
PrDiMP           & ECCV'20 & 45.12 & 75.51 & 73.72 & 65.90 & 92.58 & 91.69 & 45.05 & 54.38 & 44.09 \\
KYS              & ECCV'20 & 47.20 & 80.64 & 79.31 & 65.70 & 93.82 & 93.03 & 48.27 & 59.00 & 47.53 \\
SuperDiMP        & ECCV'20 & 49.97 & 85.24 & 83.26 & 65.74 & 94.06 & 93.34 & 45.36 & 55.46 & 44.41 \\
KeepTrack        & ECCV'21 & 49.67 & 83.69 & 81.80 & 65.04 & 90.33 & 89.40 & 40.39 & 48.81 & 48.81 \\
ToMP-101         & CVPR'22 & 55.92 & 80.01 & 79.55 & 57.79 & 78.10 & 76.92 & 39.10 & 44.41 & 44.41 \\
ToMP-50          & CVPR'22 & 58.95 & 84.37 & 84.76 & 67.40 & 91.25 & 90.55 & 40.48 & 45.63 & 39.55 \\
RTS              & ECCV'22 & 53.34 & 85.92 & 82.71 & 67.25 & 92.52 & 90.08 & 36.96 & 40.32 & 36.02 \\
DFTrack          & JSTARS'22 & 52.29 & 86.54 & 79.31 & \textbf{71.10} & 90.62 & 87.09 & 45.64 & 70.40 & 64.74 \\
TATrack          & TGRS'23 & 54.87 & 84.98 & 77.02 & 62.72 & 93.18 & 86.27 & 43.23 & 65.54 & 61.16 \\
AQAtrack-384     & CVPR'24 & 54.19 & 79.49 & 66.86 & \underline{68.94} & 94.66 & \underline{94.28} & 38.21 & 60.91 & 56.99 \\
AQAtrack-256     & CVPR'24 & 57.01 & 85.06 & 69.75 & 68.28 & \underline{94.86} & 94.04 & 41.37 & 67.58 & 63.98 \\
LoRAT-B-378      & ECCV'24 & 59.46 & 85.59 & 82.89 & 61.73 & 91.94 & 93.50 & 43.80 & 64.99 & 56.01 \\
LoRAT-L-378      & ECCV'24 & 54.26 & 87.86 & 84.15 & 52.07 & 92.64 & 93.80 & 40.20 & 68.52 & 55.94 \\
LoRAT-G-378      & ECCV'24 & 57.22 & 87.52 & 85.01 & 55.32 & 92.32 & 93.30 & 41.44 & 69.59 & 58.34 \\
\midrule
\multicolumn{11}{l}{\textit{Zero-shot methods}} \\
SAM2             & ICLR'25 & \underline{63.52} & 83.95 & 82.17 & 52.68 & 94.45 & 92.54 & \underline{48.06} & \underline{76.14} & \underline{74.60} \\
SAMURAI          & Arxiv'24 & 60.35 & \underline{88.20} & \underline{87.96} & 52.88 & 94.32 & 92.44 & 41.95 & 67.99 & 65.85 \\
\rowcolor{gray!15}
Ours             & -        & \textbf{67.23} & \textbf{92.59} & \textbf{90.58} & 68.17 & \textbf{94.95} & \textbf{94.56} & \textbf{54.15} & \textbf{78.63} & \textbf{76.37} \\
\bottomrule
\end{tabular}
\label{tab:tracker_comparison_simplified}
\end{table*}

\section{Experimental Results}
\label{sec:experiments}

\subsection{Experimental Setup}

\paragraph{Benchmarks} We evaluate SatSAM2 on four satellite video tracking benchmarks that together span the main challenges of this domain. The newly proposed \textbf{MVOT} dataset serves as our primary testbed, as it provides dense attribute annotations (illumination, viewing angle, occlusion) that enable the fine-grained analyses in Section~\ref{sec:discussion}. We additionally adopt three public benchmarks: \textbf{OOTB}, which contains 110 satellite sequences with diverse object categories; \textbf{SatSOT}, which consists of 105 sequences with 27{,}664 frames and emphasizes long-term stable tracking; and the real-world \textbf{SAT-MTB}~\cite{10130311} benchmark, which we use to assess cross-dataset generalization. All four benchmarks are evaluated under a single set of hyperparameters, so that the measured gains reflect the contribution of the framework design rather than per-dataset tuning.

\paragraph{Implementation Details} SatSAM2 is built on the pretrained \texttt{sam2.1\_hiera\_base\_plus} (b+) backbone and runs in a strict zero-shot regime with no fine-tuning or parameter updates. All inference is conducted on a single NVIDIA GeForce RTX 3090 GPU. Following SAMURAI, we set the memory affinity threshold $\tau_{\text{mask}} = 0.5$, the object existence threshold $\tau_{\text{obj}} = 0.0$, and the motion consistency threshold $\tau_{\text{kf}} = 0.0$. The Kalman motion weight is fixed at $\alpha_{\text{kf}} = 0.2$, the deformation tolerance at $\mathcal{D} = [0.5, 2.0]$, and the adaptive spatial gate at $d_{\max} = \sqrt{S}$, where $S$ denotes the initial target area. The temporal stability duration is $T_{f} = 12$ frames and the high-confidence affinity threshold is $\tau_{h} = 0.3$. A complete list of hyperparameters and their roles is provided in \cref{tab:hyperparams}; these values are kept fixed across all four benchmarks.

\paragraph{Evaluation Metrics} Following standard protocols in visual tracking, we report three metrics: the area under the success-rate curve (\textbf{AUC}), which measures overall bounding-box overlap quality; the precision at a 20-pixel center-distance threshold (\textbf{P}), which reflects localization accuracy; and the normalized precision (\textbf{$P_{\text{norm}}$}), which compensates for target scale variation across sequences.

\subsection{Comparison with State-of-the-Art Methods}

\paragraph{Main Benchmarks}
\cref{tab:tracker_comparison_simplified} compares SatSAM2 against 15 supervised trackers and two zero-shot SAM2-based baselines on OOTB, MVOT, and SatSOT. Despite operating in a strict zero-shot manner, SatSAM2 attains the best AUC on OOTB and SatSOT, and the strongest precision and normalized precision on all three benchmarks. On OOTB, SatSAM2 reaches 67.23\% AUC, improving over the strongest zero-shot baseline (SAM2, 63.52\%) by 3.71\% and over the best supervised competitor (LoRAT-B-378, 59.46\%) by 7.77\%. On SatSOT, our method surpasses the second-best tracker by 6.09\% in AUC and 2.49\% in Precision, highlighting its robustness on long, stable sequences. On MVOT, SatSAM2 achieves the highest Precision (94.95\%) and normalized Precision (94.56\%) and remains competitive in AUC against top supervised trackers, despite having never seen any remote-sensing training data. The contrast with the two SAM2-based baselines is particularly informative: both vanilla SAM2 and SAMURAI drop below 53\% AUC on MVOT, whereas SatSAM2 reaches 68.17\%. This gap confirms that generic SAM2-style trackers, which are optimized for natural video, cannot be directly transferred to satellite imagery without the motion-aware and state-aware mechanisms we propose.

\paragraph{Real-world Generalization}
To further probe cross-dataset generalization, we evaluate SatSAM2 on the real-world SAT-MTB benchmark without changing any hyperparameter. As summarized in \cref{tab:satmtb}, SatSAM2 consistently outperforms both the supervised baseline RTS and the zero-shot baselines SAM2 and SAMURAI on all three metrics, achieving a 3.43\%--3.61\% gain in AUC and a 6.59\%--6.65\% gain in Precision. This result indicates that the benefit of SatSAM2 is not tied to any particular benchmark distribution, but instead reflects a generic advantage of coupling segmentation priors with motion and state-aware reasoning.

\begin{table}[t]
\centering
\fontsize{9pt}{9pt}\selectfont
\setlength{\tabcolsep}{2mm}
\caption{\textbf{Evaluation on the SAT-MTB benchmark.} SatSAM2 demonstrates consistent superiority on an additional real-world satellite dataset.}
\begin{tabular}{l|ccc}
\toprule
\textbf{Tracker} & \textbf{AUC (\%)} & \textbf{P (\%)} & \textbf{$P_{\text{norm}}$ (\%)} \\
\midrule
RTS & 50.74 & 58.22 & 50.23 \\
SAM2 & 51.11 & 59.57 & 50.79 \\
SAMURAI & 51.29 & 59.51 & 51.01 \\
\rowcolor{gray!15}
\textbf{SatSAM2 (Ours)} & \textbf{54.72} & \textbf{66.16} & \textbf{54.48} \\
\bottomrule
\end{tabular}
\label{tab:satmtb}
\end{table}

\subsection{In-depth Analysis}

\begin{figure}[t]
  \centering
  \includegraphics[width=\columnwidth]{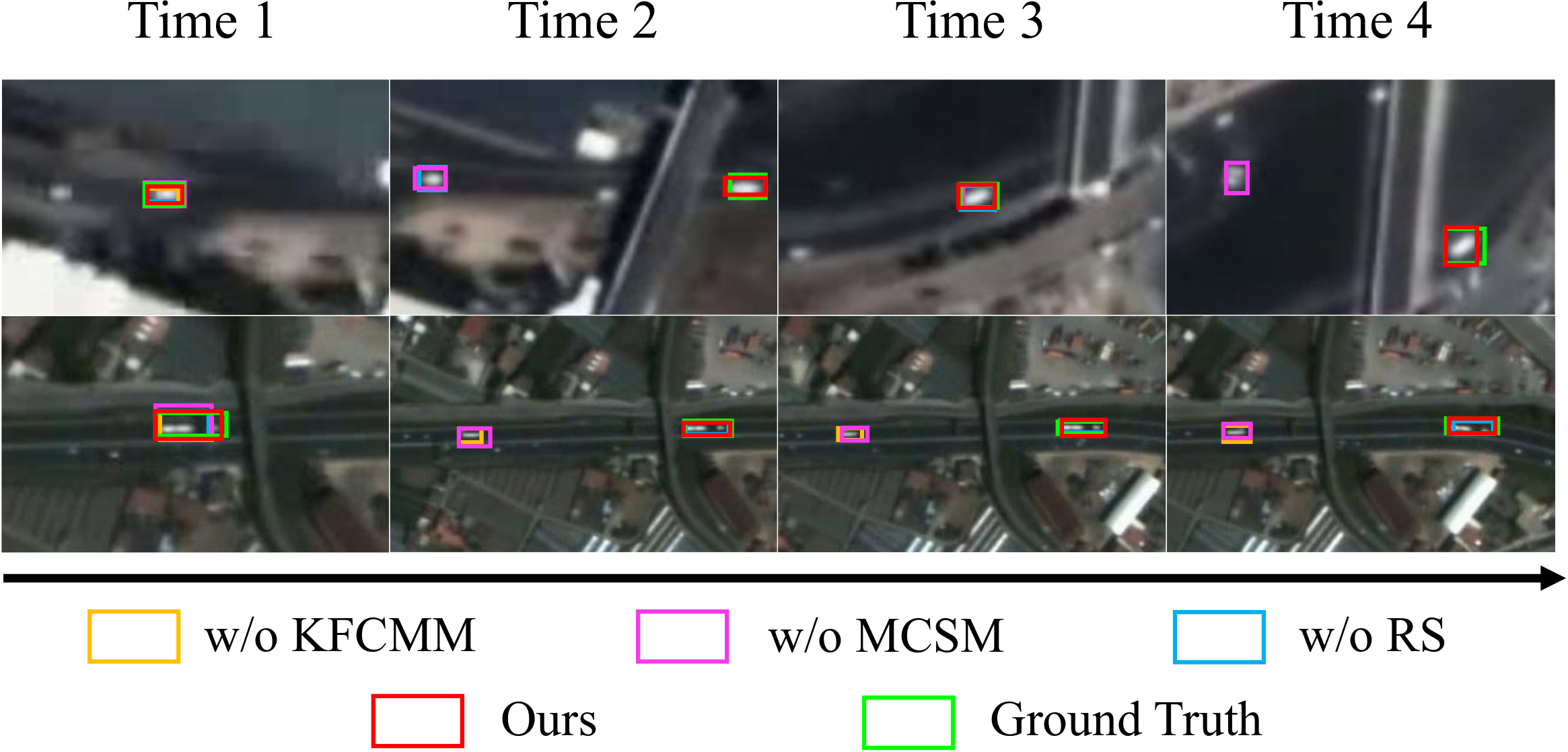}
  \caption{\textbf{Ablation comparison between Ours, GT, and variants with different modules removed.}}
  \label{fig:ablation}
\end{figure}

\paragraph{Ablation Study}
We conduct a component-level ablation on OOTB to quantify the contribution of the KFCMM motion module, the MCSM state machine, and the ResetStable (RS) recovery mechanism. Quantitative results are reported in \cref{tab:ablation} and visualized in \cref{fig:ablation}. Removing KFCMM causes the most pronounced drop (67.23\% $\rightarrow$ 61.01\% AUC), confirming that a Kalman-based motion prior is essential for disambiguating similar targets and for keeping the tracker anchored during temporary appearance degradation. Disabling MCSM also degrades AUC to 62.71\%, showing that an explicit state machine is needed to decide when the motion prior should dominate over visual evidence. Removing RS causes a smaller but still consistent drop (66.95\%), indicating that the recovery mechanism is mainly responsible for correcting residual drift in difficult sequences. A particularly telling observation is that the \textbf{Naive-SAM2+KF} baseline, which naively attaches a standard Kalman filter to SAM2 without our KFCMM constraints or MCSM logic, collapses to 36.95\% AUC, well below both SAM2 and SAMURAI. This negative result confirms that simply concatenating a motion filter to a foundation segmenter is insufficient; the gain of SatSAM2 comes from the co-design of constrained motion modeling, state-aware selection, and recovery, not from the Kalman filter alone.

\begin{table}[t]
\centering
\caption{\textbf{Quantitative ablation results on the OOTB dataset.} Naive-SAM2+KF denotes SAM2 with a standard Kalman filter but without KFCMM constraints or MCSM state management.}
\label{tab:ablation}
\fontsize{9pt}{9pt}\selectfont
\begin{tabular}{l|c|c|c}
\toprule
\textbf{Tracker} & \textbf{AUC (\%)} & \textbf{P (\%)} & \textbf{$P_{\text{norm}}$ (\%)} \\
\midrule
Baseline (SAM2) & 63.52 & 83.95 & 82.17 \\
Baseline (SAMURAI) & 60.35 & 88.20 & 87.96 \\
Naive-SAM2+KF & 36.95 & 50.96 & 51.08 \\
\midrule
Ours (w/o KFCMM) & 61.01 & 81.79 & 80.79 \\
Ours (w/o MCSM) & 62.71 & 85.26 & 82.38 \\
Ours (w/o RS) & \underline{66.95} & \underline{90.67} & \underline{88.23} \\
\rowcolor{gray!15}
Ours & \textbf{67.23} & \textbf{92.59} & \textbf{90.58} \\
\bottomrule
\end{tabular}
\end{table}

\begin{figure*}[t]
\centering
\includegraphics[width=\textwidth]{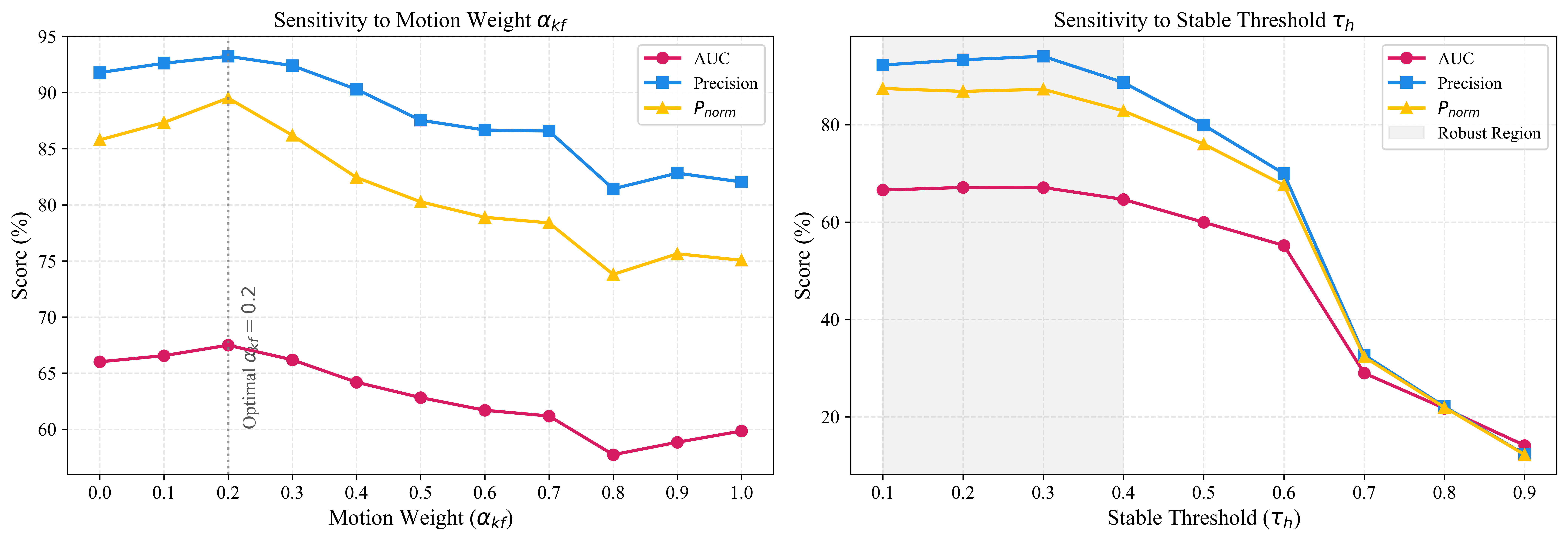}
\caption{\textbf{Hyperparameter sensitivity analysis on the OOTB dataset.}
\textbf{(Left)}~Impact of the Kalman motion weight ($\alpha_{kf}$): performance peaks at $\alpha_{kf}=0.2$, confirming that a subtle motion constraint effectively complements visual features.
\textbf{(Right)}~Impact of the stability threshold ($\tau_{h}$): the method is robust across $\tau_{h} \in [0.1, 0.4]$. Overly strict criteria ($\tau_{h} \ge 0.6$) prevent stable-state transitions and degrade performance.}
\label{fig:sensitivity}
\end{figure*}

\paragraph{Hyperparameter Sensitivity}
Beyond verifying \emph{what} each module contributes, we also study \emph{how sensitive} the framework is to its two most critical hyperparameters: the Kalman motion weight $\alpha_{kf}$ and the high-confidence stability threshold $\tau_h$. \cref{fig:sensitivity} reports the performance of SatSAM2 on OOTB as each parameter is varied in isolation while all others remain fixed. For the motion weight (\cref{fig:sensitivity}, left), the framework exhibits a clear robustness zone for $\alpha_{kf} \in [0.0, 0.3]$, within which the tracker leverages the motion prior to resolve visual ambiguities without overriding valid segmentation evidence. Performance peaks at $\alpha_{kf} = 0.2$ and then degrades monotonically as $\alpha_{kf}$ exceeds $0.4$, where the linear motion assumption begins to dominate and suppress useful appearance cues. For the stability threshold (\cref{fig:sensitivity}, right), SatSAM2 shows a wide plateau over $\tau_h \in [0.1, 0.4]$ and peaks at $\tau_h = 0.3$. When $\tau_h \geq 0.6$, the state machine becomes overly conservative: the \textit{Stable} transition is rarely triggered, the advanced decision logic of MCSM is effectively disabled, and performance regresses toward the baseline. These two curves together indicate that SatSAM2 is not the product of delicate tuning, as the operating point used throughout the paper lies near the center of a broad stable region for both parameters.

\begin{figure}[t]
\centering
\includegraphics[width=\linewidth]{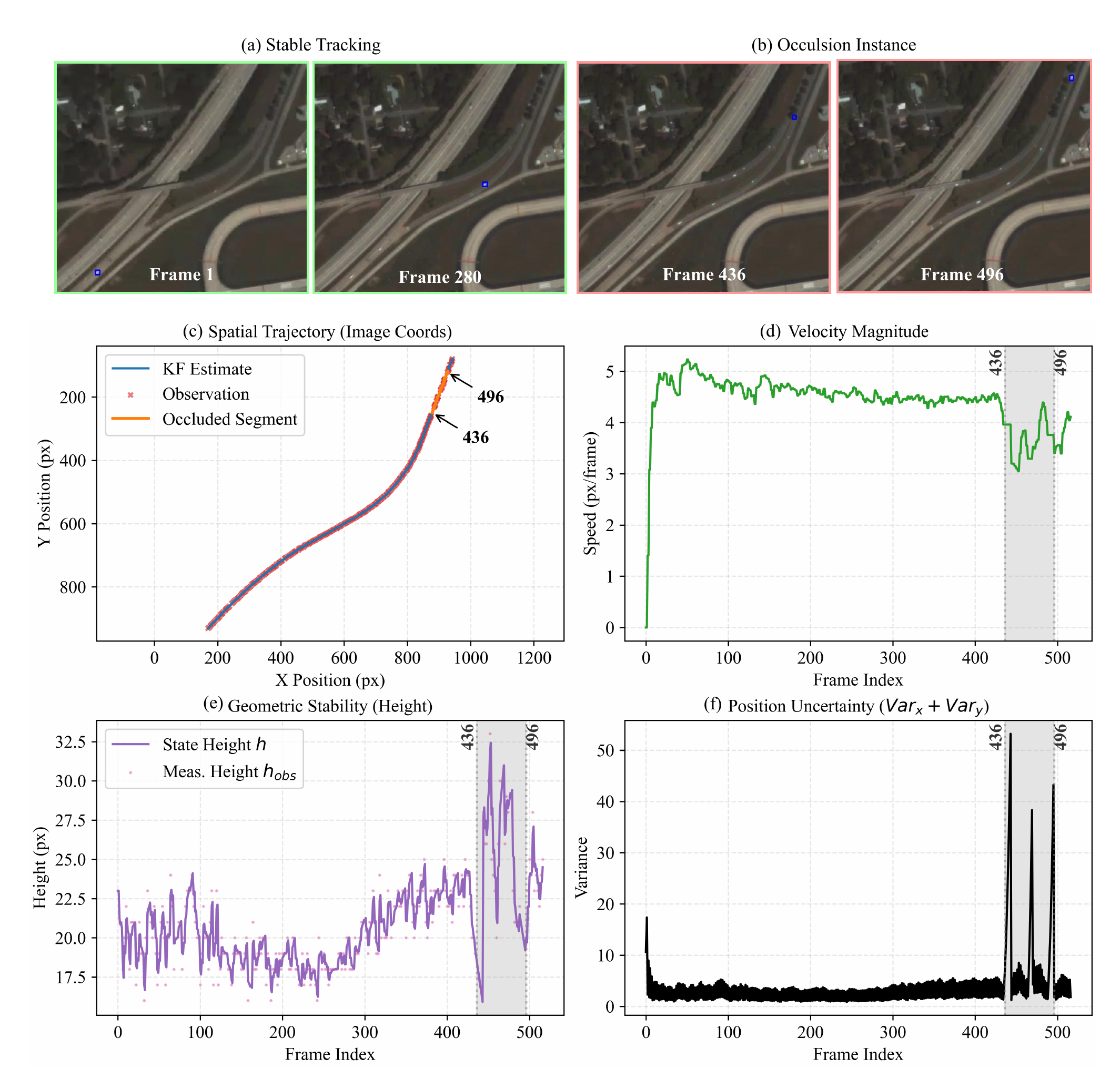}
\caption{\textbf{Visualization of Kalman filter state dynamics on sequence car\_30.}
(a-b)~Qualitative results: stable tracking (a) and a severe occlusion event (b) in which the target is obscured by trees (frames 436--496).
(c)~Spatial trajectory: the estimated path (blue) remains smooth during the occlusion, bridging the gap in visual observations (red crosses).
(d)~Velocity magnitude: the filter maintains a consistent velocity prior during signal loss.
(e)~Geometric stability: the state height $h$ (purple) remains stable despite missing measurements, validating the rigid-body constraint.
(f)~Position uncertainty: the aggregated variance $(\mathrm{Var}_x + \mathrm{Var}_y)$ spikes during occlusion, triggering the MCSM to prioritize motion priors.}
\label{fig:state_dynamics}
\end{figure}

\paragraph{Internal State Dynamics of KFCMM}
To verify that the KFCMM behaves as intended and to expose the internal mechanism behind occlusion recovery, we visualize its state evolution on the challenging sequence \texttt{car\_30}, in which the target is completely occluded by roadside vegetation between frames 436 and 496 (\cref{fig:state_dynamics}a--b). \cref{fig:state_dynamics}f shows that the aggregated position uncertainty $(\mathrm{Var}_x + \mathrm{Var}_y)$ grows rapidly as visual observations become unreliable; this uncertainty spike is used by the MCSM as an explicit signal to switch from the \textit{Update} phase to a pure \textit{Prediction} phase. During this prediction phase, the tracker extrapolates the trajectory using the velocity state accumulated before occlusion (\cref{fig:state_dynamics}d). The resulting spatial path (\cref{fig:state_dynamics}c) remains smooth and aligns with the road geometry, so that the tracker can immediately re-associate with the target once it re-emerges. Equally important, \cref{fig:state_dynamics}e shows that the estimated state height $h$ remains stable throughout the occluded interval. In contrast, the raw SAM2 segmentation mask can collapse or expand under partial occlusion, but the geometric constraint of KFCMM prevents this ``box collapse'' behavior. Together, these state-space diagnostics provide an interpretable explanation for the strong occlusion robustness observed quantitatively both on the main benchmarks and in the attribute-level analysis of Section~\ref{sec:discussion}.

\paragraph{Computational Efficiency}
Finally, we evaluate the inference speed (FPS) and peak GPU memory usage of SatSAM2 on SatSOT using a single NVIDIA RTX 3090, and compare against both lightweight supervised trackers and the SAM2 family (\cref{tab:efficiency}). As expected, segmentation-based trackers are slower than feature-correlation trackers such as DiMP-50. More importantly, the additional KFCMM and MCSM modules introduce only negligible overhead relative to vanilla SAM2: SatSAM2 runs at 14.45 FPS (vs.\ 14.84 FPS for SAM2), a slowdown of merely 0.39 FPS, and consumes only 0.027 GB of additional GPU memory. Since commercial satellite video feeds typically operate between 10 and 25 FPS, SatSAM2 comfortably satisfies real-time constraints while providing substantially better accuracy.

\begin{table}[t]
\centering
\fontsize{9pt}{9pt}\selectfont
\setlength{\tabcolsep}{3mm}
\caption{\textbf{Computational efficiency comparison} on SatSOT (single NVIDIA RTX 3090).}
\begin{tabular}{l|cc}
\toprule
\textbf{Method} & \textbf{FPS} & \textbf{Mem (GB)} \\
\midrule
DiMP-50 & 36.08 & 0.526 \\
ToMP-50 & 24.93 & 0.392 \\
KeepTrack & 14.82 & 0.982 \\
\midrule
SAM2 & 14.84 & 1.524 \\
SAMURAI & 14.53 & 1.536 \\
\rowcolor{gray!15}
\textbf{SatSAM2 (Ours)} & 14.45 & 1.551 \\
\bottomrule
\end{tabular}
\label{tab:efficiency}
\end{table}

\subsection{Qualitative Evaluation}
\begin{figure*}[t]
\centering
\includegraphics[width=\linewidth]{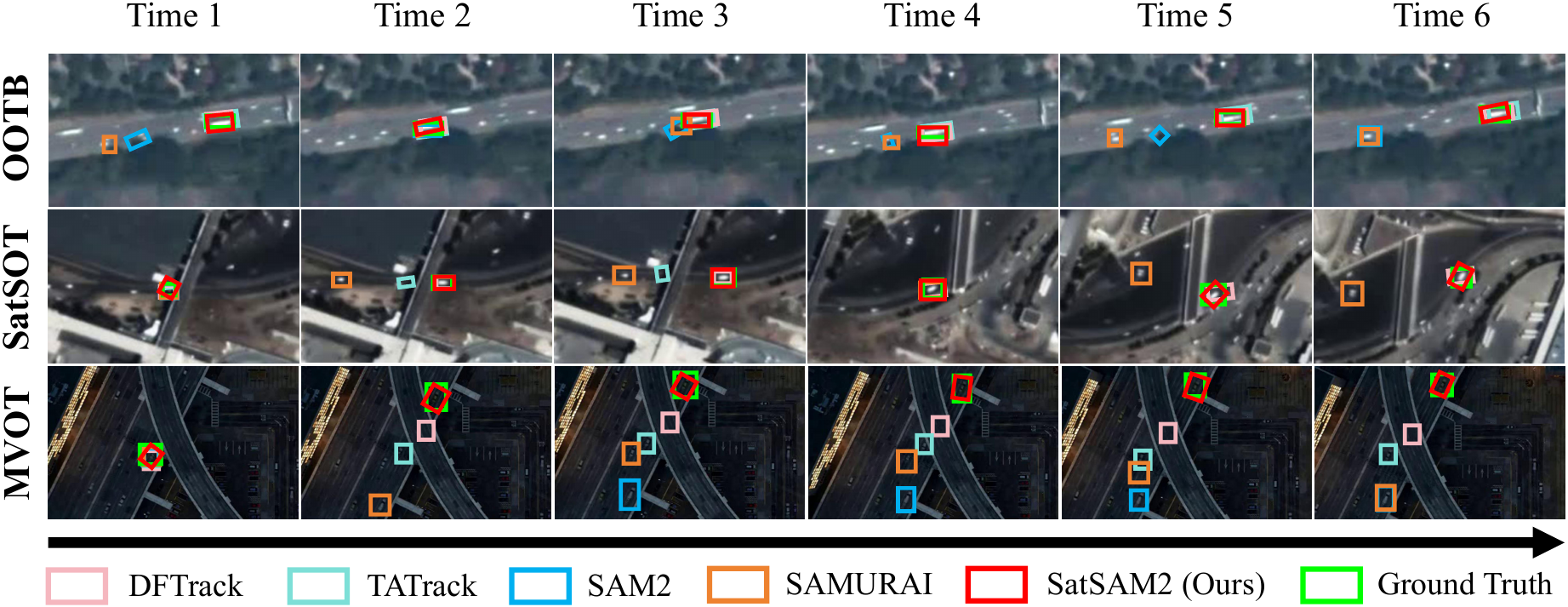}
\caption{\textbf{Qualitative comparison on three remote sensing datasets.} Our method achieves the most accurate tracking results and demonstrates strong robustness in recovering targets after occlusion, enabling fast and reliable re-alignment.}
\label{fig:qualitative}
\end{figure*}

\begin{figure}[t]
\centering
\includegraphics[width=\linewidth]{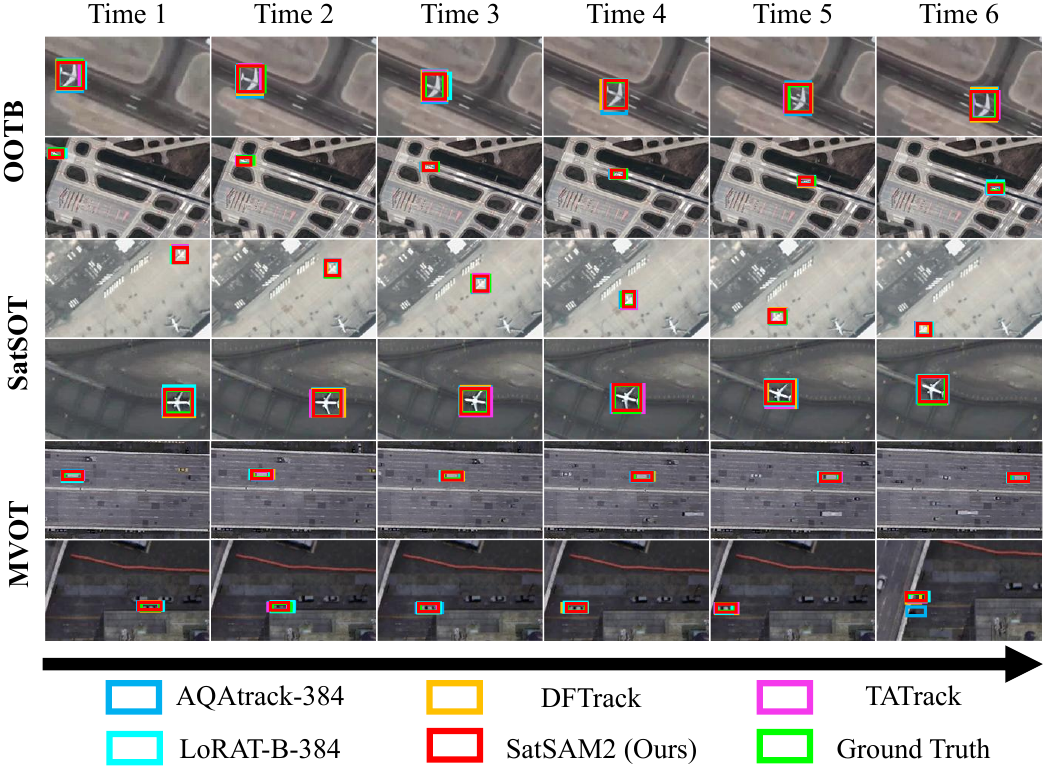}
\caption{Qualitative comparisons on three datasets among our method and fully supervised methods (AQAtrack and LoRAT with different parameters). Each row represents a different dataset, and each column shows tracking results from a different method. Our method demonstrates superior robustness and temporal consistency.}
\label{fig:fldx}
\end{figure}

\cref{fig:qualitative} presents qualitative comparisons on representative sequences that contain multiple visually similar targets and temporary occlusions. SatSAM2 maintains correct identity even when several distractors enter the search region, and it cleanly re-acquires the target after brief occlusion by bridges or elevated structures. We directly attribute this behavior to the motion-informed selection logic analyzed in \cref{fig:state_dynamics}. \cref{fig:fldx} further contrasts SatSAM2 with representative supervised trackers (AQAtrack, LoRAT) and domain-specific remote-sensing trackers (DFTrack, TATrack) on simpler sequences from three datasets~\cite{9672083, CHEN2024212}. Across these visualizations, our predicted bounding boxes align more tightly with the ground truth and exhibit substantially less temporal jitter, which is consistent with the higher precision and normalized precision reported in \cref{tab:tracker_comparison_simplified}.

\section{Discussion}
\label{sec:discussion}
Section~\ref{sec:experiments} has established that SatSAM2 achieves state-of-the-art tracking accuracy on four satellite benchmarks and has dissected the role of each module through component-level and state-level analyses. In this section, we take a complementary perspective and examine \emph{when and why} the method works, by leveraging the dense attribute annotations of the MVOT dataset. We first analyze tracking behavior under three key environmental conditions, namely illumination, viewing angle, and occlusion, which jointly characterize the operational envelope of satellite video. We then discuss representative failure modes and conclude with a reflection on the broader applicability of the proposed design.

\subsection{Attribute-based Analysis on MVOT}
\label{sec:attribute_analysis}

A distinguishing property of MVOT is that every sequence is annotated with illumination, viewing-angle, and occlusion attributes. This allows us to disentangle the contribution of environmental conditions from object-level difficulty, and to examine how different families of trackers respond to each factor.

\subsubsection{Impact of Illumination Conditions}
\label{sec:illumination}

\begin{figure}[t]
\centering
\includegraphics[width=\linewidth]{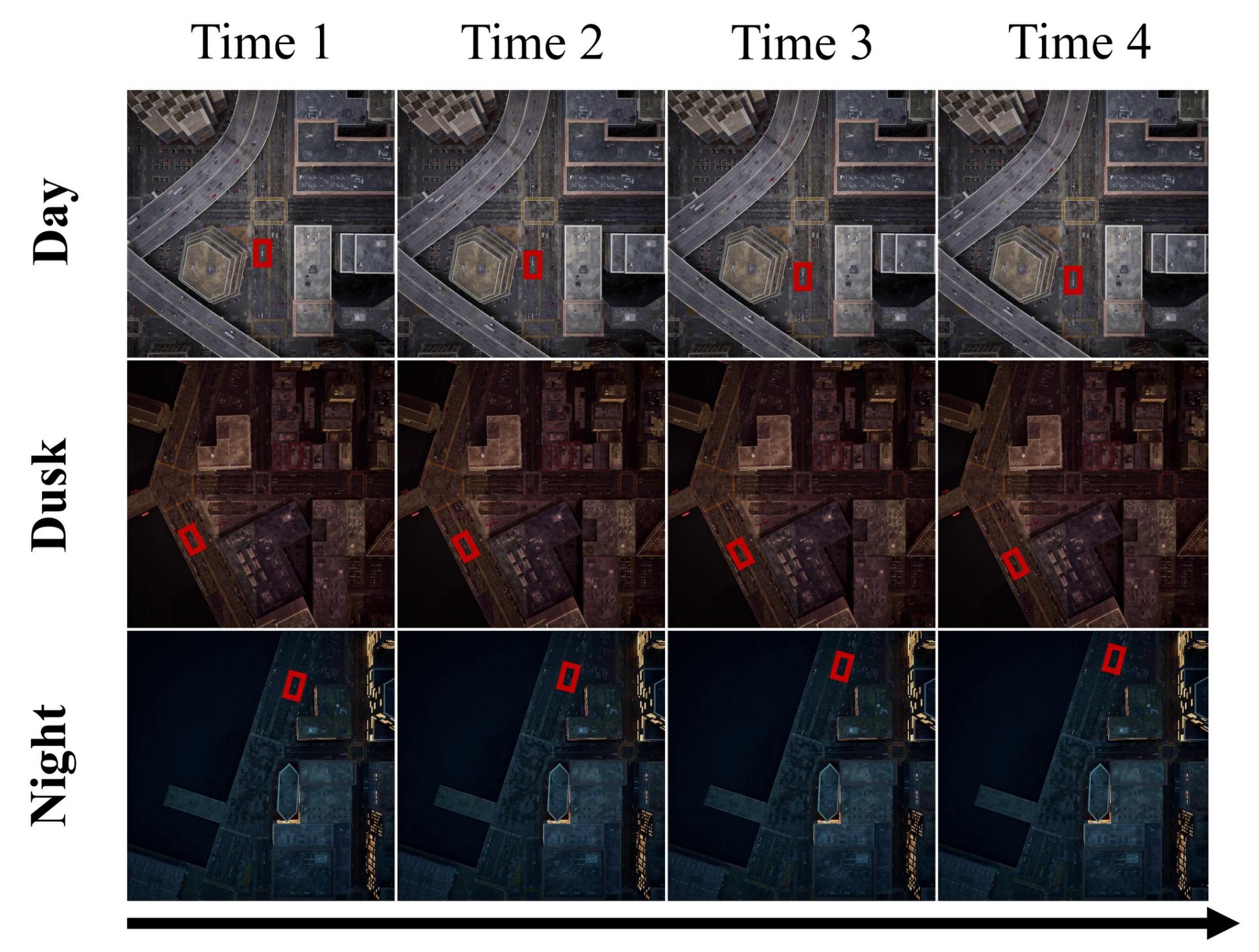}
\caption{\textbf{Example scenes under different illumination conditions (Day, Dusk, and Night).} The red boxes indicate the target locations in different frames.}
\label{fig:day}
\end{figure}
MVOT comprises 94,200 daytime, 31,600 dusk, and 32,100 nighttime frames, and thus covers a wide range of illumination conditions with markedly different visual characteristics (\cref{fig:day}). Across all methods, performance consistently degrades at night, which is expected given reduced visibility and weaker appearance cues. Zero-shot segmentation-based approaches, including SAM2, SAMURAI, and our method, achieve slightly lower AUC than supervised trackers in low-light scenes while maintaining comparable precision. We attribute this pattern to the tendency of segmentation-based methods to overestimate mask sizes under ambiguous lighting, which inflates the bounding box without necessarily moving its center. Despite this challenge, SatSAM2 still delivers the best overall performance across all three illumination regimes, and its advantage is particularly pronounced during the day (AUC 73.22\%, P 96.98\%), where appearance cues are most reliable and the motion prior acts as a stabilizing rather than compensating force.

\subsubsection{Impact of Viewing Angles}
\label{sec:viewing_angles}

\begin{figure}[t]
\centering
\includegraphics[width=\linewidth]{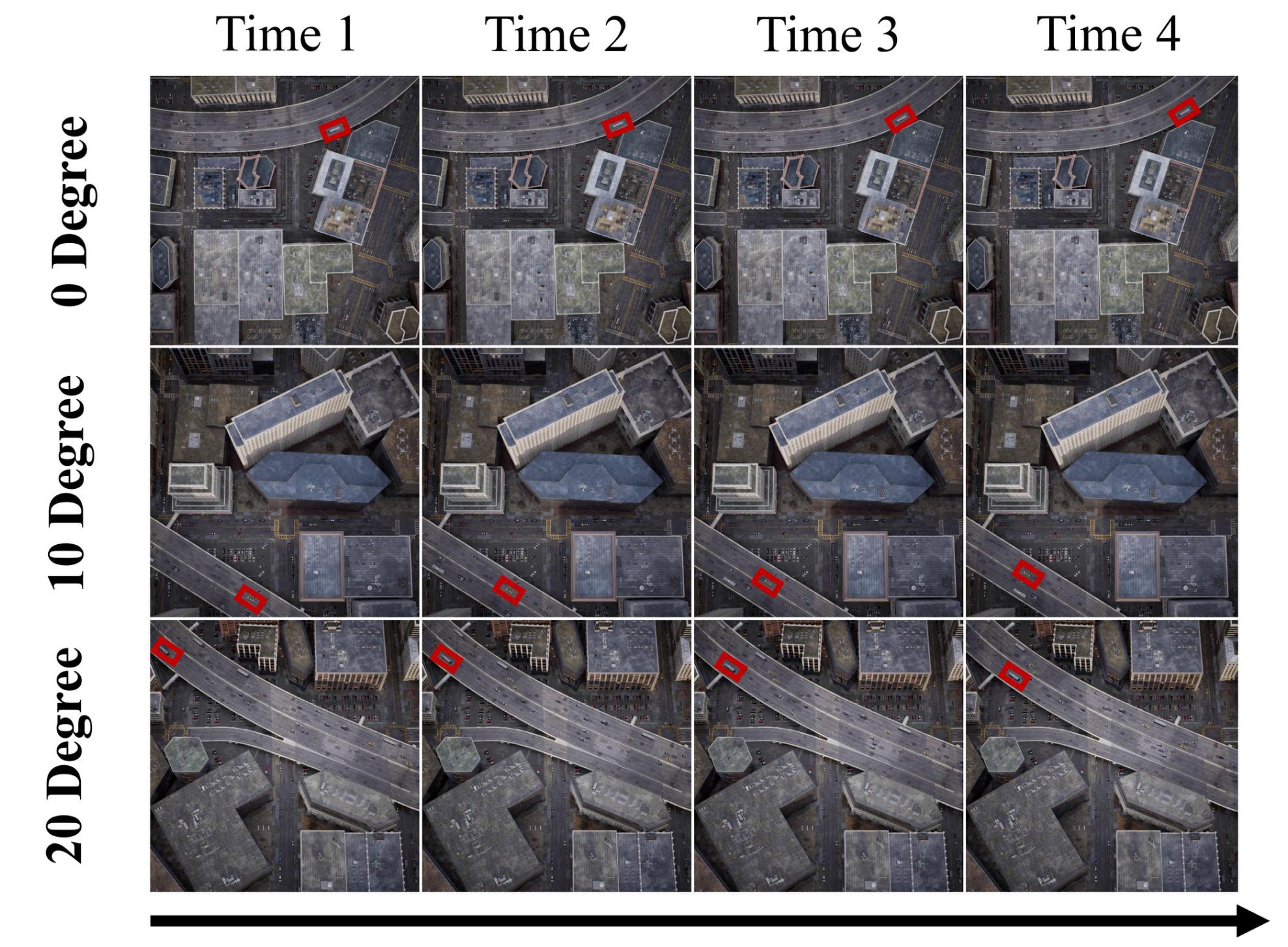}
\caption{\textbf{Example scenes under different viewing angles (0\textdegree, 10\textdegree, and 20\textdegree).} Red boxes indicate the target positions in different frames.}
\label{fig:degree}
\end{figure}

MVOT contains \textbf{91{,}800 frames} at a 0$^\circ$ viewing angle, \textbf{31{,}400 frames} at 10$^\circ$, and \textbf{31{,}000 frames} at 20$^\circ$; representative examples are shown in \cref{fig:degree}. We observe that performance peaks at 10$^\circ$ and is lowest at 0$^\circ$. The purely top-down view (0$^\circ$) provides essentially no lateral information, forcing the model to rely on a compressed silhouette that easily blends into the background during fast motion. The 20$^\circ$ viewing angle, while richer in appearance cues, introduces a higher rate of occlusions caused by tall buildings and other structures. A moderate 10$^\circ$ angle strikes a balance between geometric detail and unobstructed visibility, which explains the peak performance. SatSAM2 consistently outperforms all competitors in precision and normalized precision across the three viewing angles, indicating that the motion prior compensates for the appearance degradation typical of bird's-eye views.

\subsubsection{Robustness under Occlusion}
\label{sec:occlusion_analysis}

\begin{figure}[t]
\centering
\includegraphics[width=\linewidth]{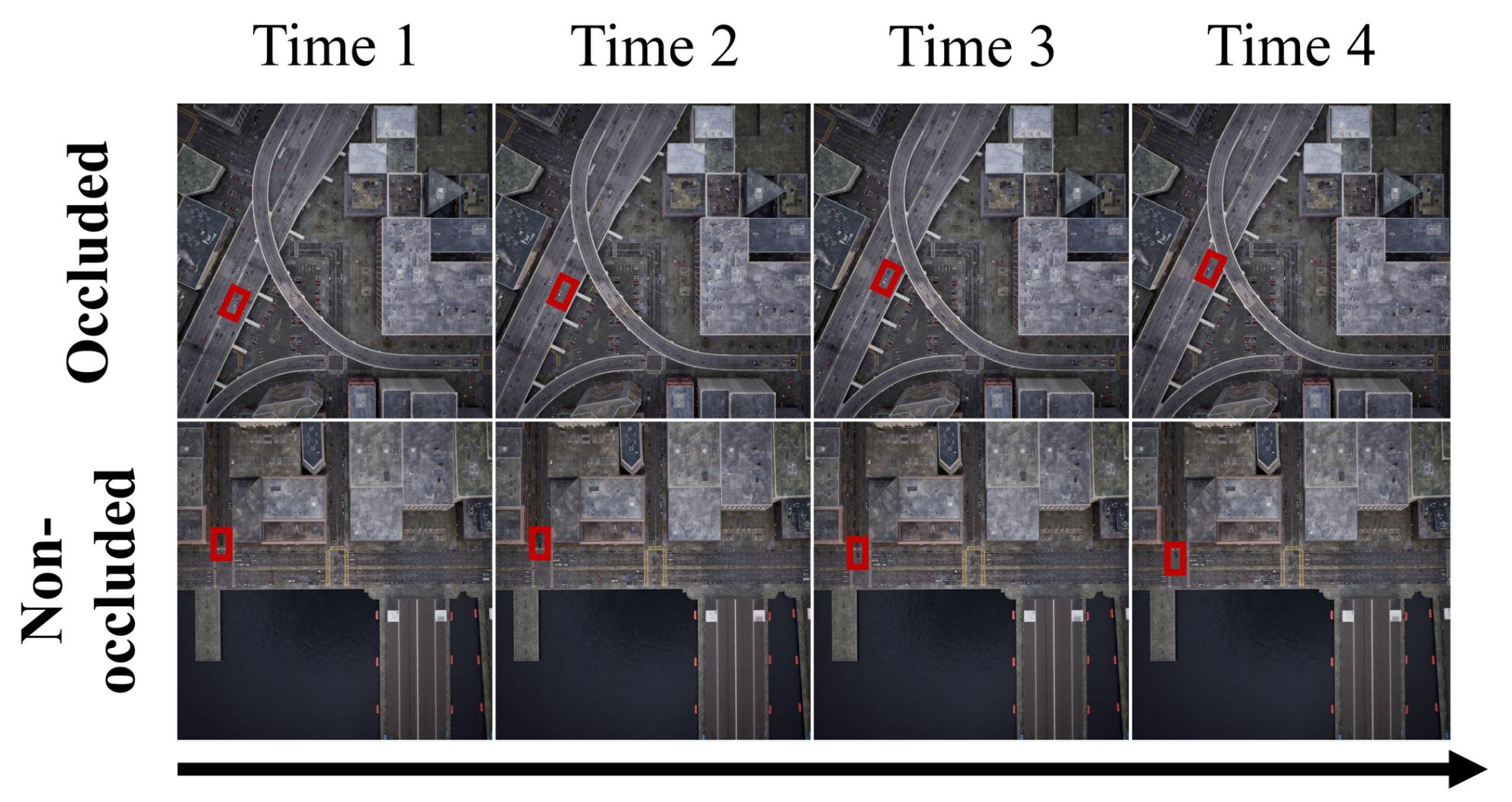}
\caption{\textbf{Example scenes under occluded and non-occluded conditions.} Red boxes indicate the target positions in different frames.}
\label{fig:occ}
\end{figure}

Occlusion is arguably the most demanding attribute in satellite tracking, since lost targets cannot be recovered from appearance alone. MVOT includes \textbf{148{,}700 non-occluded} and \textbf{9{,}200 occluded} frames, with occlusions primarily caused by elevated structures and slanted buildings (\cref{fig:occ}). \cref{tab:occlusion} reports a detailed performance breakdown: traditional trackers suffer severe degradation under occlusion (AUC dropping into the 30\% range) because they lack an explicit mechanism to predict object position once appearance evidence vanishes. In contrast, SatSAM2 achieves 44.50\% AUC under occlusion, exceeding the second-best method by 6.03\%. This improvement mirrors the internal state dynamics reported in \cref{fig:state_dynamics}: the rising positional variance triggers a transition to a pure prediction phase, and the accumulated velocity state carries the tracker through the occlusion gap until visual evidence returns. Notably, on non-occluded frames the gap between zero-shot and supervised methods narrows considerably, confirming that the principal advantage of SatSAM2 arises precisely in the regime where purely appearance-based trackers are structurally weakest.

To further justify the use of MVOT as a stress-test for occlusion robustness despite its 100-frame sequence length, we compare its occlusion statistics against two widely used real-world benchmarks in \cref{tab:long_term}. MVOT exhibits an average number of occluded frames per event (ANOF) and average target displacement (ATD) that are comparable to OOTB and SatSOT, while containing substantially more occlusion events and a higher overall occlusion rate (28.33\%). In other words, MVOT compresses a similar per-event difficulty into a denser and more frequent occurrence, which we regard as more informative for evaluating recovery mechanisms. Future releases of MVOT will additionally include longer single-sequence videos to better support long-term tracking evaluation.

\begin{table}[t]
\centering
\fontsize{9pt}{9pt}\selectfont
\setlength{\tabcolsep}{2mm}
\caption{\textbf{Occlusion and displacement statistics across benchmarks.} MVOT exhibits comparable or more challenging occlusion characteristics relative to real-world datasets.}
\begin{tabular}{l|cccc}
\toprule
\textbf{Dataset} & \textbf{Occ. Events} & \textbf{ANOF} & \textbf{Occ. Rate} & \textbf{ATD (px)} \\
\midrule
OOTB & 11 & 17.4 & 5.73\% & 303.76 \\
SatSOT & 21 & 28.7 & 13.11\% & 294.85 \\
\rowcolor{gray!15}
\textbf{MVOT-Occ.} & \textbf{109} & \textbf{23.7} & \textbf{28.33\%} & \textbf{264.02} \\
\bottomrule
\end{tabular}
\label{tab:long_term}
\end{table}

\begin{table}[t]
\centering
\setlength{\tabcolsep}{1mm}
\fontsize{9pt}{9pt}\selectfont
\caption{Tracking performance under occlusion and non-occlusion. Best results are bolded, second best underlined.}
\begin{tabular}{l|ccc|ccc}
\toprule
\multirow{2}{*}{\textbf{Tracker}} & \multicolumn{3}{c|}{\textbf{Occlusion}} & \multicolumn{3}{c}{\textbf{Non-Occlusion}} \\
 & \textbf{AUC} & \textbf{P} & \textbf{$P_{\text{norm}}$} & \textbf{AUC} & \textbf{P} & \textbf{$P_{\text{norm}}$} \\
\midrule
\multicolumn{7}{l}{\textit{Supervised methods}} \\
ToMP-50 & 37.80 & 51.79 & 52.54 & 70.85 & 94.50 & 95.23 \\
ToMP-101 & 33.00 & 45.63 & 46.60 & 57.22 & 76.06 & 77.64 \\
DiMP-50 & 34.77 & 51.96 & 53.68 & 65.37 & 95.09 & 96.65 \\
SuperDiMP & 34.84 & 50.53 & 51.61 & 70.50 & 95.84 & 96.79 \\
PrDiMP & 36.08 & 51.94 & 53.21 & 70.63 & 94.45 & 95.33 \\
RTS-50 & 34.69 & 50.00 & 51.77 & 70.47 & 95.33 & 96.11 \\
KeepTrack & 32.77 & 46.16 & 47.39 & 70.87 & 93.72 & 94.61 \\
ATOM & 34.56 & 49.88 & 51.72 &  \underline{71.51} & 95.85 & 96.61 \\
DFTrack  & 32.37 & 48.52 & 45.44 & \textbf{72.87} & 92.55 & 88.94 \\
TATrack  & 33.12 & 50.29 & 47.68 & 64.08 & 95.15 & 87.97  \\
AQAtrack-256 & 37.74 & 36.55 & 46.59 & 69.77 & 95.05 & 95.47 \\
AQAtrack-384 &  \underline{38.47} & 37.28 & 46.71 & 70.41 & 94.89 & 95.33 \\
LoRAT-G-378 & 32.57 & 53.72 & 54.21 & 58.53 & 96.40 & 96.81 \\
LoRAT-L-378 & 30.12 & 54.17 & 54.61 & 54.96 & \underline{96.73} & \textbf{97.14} \\
LoRAT-B-378 & 36.19 & 54.14 & 54.50 & 65.60 & 96.00 & 96.40 \\
\midrule
\multicolumn{7}{l}{\textit{Zero-shot methods}} \\
SAM2 & 31.07 & \underline{56.18} & \underline{57.04} & 62.43 & 93.20 & 94.44 \\
SAMURAI & 30.66 & 56.00 & 56.48 & 63.02 & 94.13 & 95.31 \\
\rowcolor{gray!15}
Ours & \textbf{44.50} & \textbf{59.03} & \textbf{60.35} & 68.67 & \textbf{97.00} & \underline{97.10} \\
\bottomrule
\end{tabular}
\label{tab:occlusion}
\end{table}

\subsection{Failure Case Analysis}
\label{sec:limits}

\begin{figure}[t]
\centering
\includegraphics[width=\linewidth]{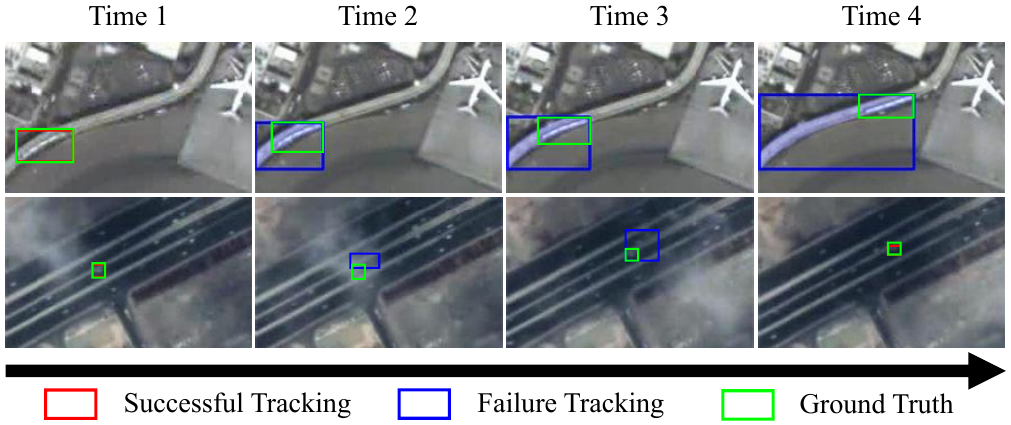}
\caption{\textbf{Qualitative analysis of failure cases.}
\textbf{(Top Row)}~Gradual drift on elongated targets: SAM2 over-segmentation evolves smoothly, evading the Kalman filter's outlier rejection.
\textbf{(Bottom Row)}~Low-contrast ambiguity: the lack of distinct boundary features causes the segmentation mask to drift into the background.}
\label{fig:failure_cases}
\end{figure}

While the preceding analyses show that SatSAM2 performs robustly across a wide range of operating conditions, a principled evaluation also requires understanding where the framework breaks down. \cref{fig:failure_cases} illustrates the two most representative failure modes, which we discuss below together with two additional limitations inherited from the underlying components.

\noindent\textbf{Boundary leakage on slender targets.}
A distinct failure mode occurs when tracking elongated objects such as trains, particularly in low-resolution imagery where edge gradients are weak. In these scenarios the SAM2 backbone tends to generate over-expanded segmentation masks. Because this mask inflation evolves smoothly in time, lacking the abrupt ``jumps'' that characterize typical tracking errors, the Kalman filter interprets the gradual expansion as valid object motion rather than as an outlier. The error thus stays inside the acceptance gate of KFCMM and produces progressive drift.

\noindent\textbf{Visual ambiguity in low-contrast regions.}
A second challenge arises from the inherent limitations of optical satellite sensors in low-contrast or shadowed regions. When a target enters a deep shadow or crosses a background with nearly identical texture, gradient boundaries effectively vanish; the SAM2 decoder then produces imprecise masks and the bounding box drifts into the background before the state machine can detect the anomaly.

\noindent\textbf{Non-linear motion under occlusion.}
SatSAM2 relies on a linear Kalman filter as its motion prior. Targets that undergo non-linear motion (e.g., a turning vehicle) \emph{while} being occluded may therefore deviate significantly from the predicted trajectory, so that when the target re-emerges the predicted position no longer falls within the spatial gate.

\noindent\textbf{Mask precision limitations.}
Because SatSAM2 uses SAM2 as its segmentation backbone, it inherits the remaining limitations of that model. In particular, the predicted masks may systematically over- or under-estimate target size, and these biases can accumulate over long sequences even when the tracker remains on target.

\subsection{Broader Applicability and Future Work}
\label{sec:broader}

Although SatSAM2 is specifically designed for satellite video, the three core ideas it builds upon are not inherently tied to the satellite domain: combining a foundation segmentation model with a constrained motion prior, using an explicit state machine to arbitrate between appearance and motion, and providing an explicit recovery mechanism. The framework could be adapted to other top-down or near-top-down scenarios, such as UAV surveillance or traffic monitoring, by relaxing the scale-invariance assumption that is appropriate for fixed-altitude satellites but not for altitude-varying platforms.

Future work will address the failure modes identified above along several directions. First, we plan to incorporate more discriminative, instance-specific appearance embeddings to reduce boundary leakage on slender targets. Second, we will investigate non-linear motion models, including extended or unscented Kalman filters and lightweight particle filters, that can better cope with maneuvering targets during occlusion. Third, we will explore multi-hypothesis tracking strategies to handle simultaneous distractors with physically plausible trajectories. Finally, extending MVOT with longer single-sequence videos will enable the community to jointly study long-term tracking, state-aware reasoning, and recovery in a unified setting.
\section{Conclusion}

In this work, we present \textbf{SatSAM2}, a motion-constrained framework that integrates the promptable segmentation power of SAM2 into the challenging task of satellite video object tracking. By combining a Kalman filter-based state machine with satellite-specific motion priors, SatSAM2 effectively addresses occlusion, appearance ambiguity, and tracking drift. Extensive experiments on multiple real-world and synthetic benchmarks demonstrate the robustness and generalization of SatSAM2, setting a new baseline for remote sensing video tracking. In future work, we plan to explore transformer-based temporal models for enhanced long-term dependency modeling and extend SatSAM2 to multi-object tracking in complex satellite scenes.

\bibliographystyle{IEEEtran}
\bibliography{main}

\end{document}